\documentclass[runningheads]{llncs}

\usepackage{eccv}

\usepackage{eccvabbrv}

\usepackage{graphicx}
\usepackage{booktabs}

\usepackage[accsupp]{axessibility}  

\usepackage{hyperref}

\usepackage{orcidlink}

\usepackage{amsmath}
\usepackage{booktabs}
\usepackage{colortbl}
\usepackage{dsfont}
\usepackage{algorithm}
\usepackage{algorithmic}
\usepackage{wrapfig}

\newcommand{\bI}{\mathbf{I}}
\newcommand{\bp}{\mathbf{p}}
\newcommand{\bx}{\mathbf{x}}

\newcommand{\bSigma}{\boldsymbol{\Sigma}}

\newcommand{\cG}{\mathcal{G}}

\newcommand{\cX}{\mathcal{X}}
\newcommand{\cT}{\mathcal{T}}
\newcommand{\cP}{\mathcal{P}}

\definecolor{tablered}{rgb}{1, 0.7, 0.7}
\definecolor{tableyellow}{rgb}{1, 1, 0.7}
\definecolor{tableorange}{rgb}{1, 0.85, 0.7}


\usepackage[capitalize]{cleveref}

\crefname{section}{Sec.}{Secs.}
\Crefname{section}{Section}{Sections}

\crefname{table}{Tab.}{Tabs.}
\Crefname{table}{Table}{Tables}

\crefname{figure}{Fig.}{Figs.}
\Crefname{figure}{Figure}{Figures}

\crefname{equation}{Eq.}{Eqs.}
\Crefname{equation}{Equation}{Equations}

\crefname{algorithm}{Alg.}{Algs.}
\Crefname{algorithm}{Algorithm}{Algorithms}

\usepackage{multirow}

\begin{document}

\title{Towards Alias-Free 4D Gaussian Representations with Motion-Aware Filtering} 

\titlerunning{Towards Alias-Free 4D Gaussian Representations}

\author{Ankit Dhiman\inst{1,2} \and
Kunal A Kathare\inst{1} \and
Pranav Vignesh\inst{1} \and
Lokesh R Boregowda\inst{2} \and
 Venkatesh Babu Radhakrishnan\inst{1}}

\authorrunning{A.~Dhiman et al.}

\institute{Indian Institute of Science, Bangalore \and
Samsung R\&D Institute India - Bangalore \\
\url{https://maaf-4dgs.github.io/}}

\maketitle

\begin{abstract}
Novel-view synthesis of dynamic scenes, crucial for AR/VR applications, remains a challenging problem. Recent methods adapt representations like 3D Gaussian Splatting (3DGS) and Neural Radiance Fields (NeRF) for dynamic scenes by incorporating time as the fourth dimension (4D representations). These 4D representations still suffer from aliasing artifacts, especially when generating novel views from divergent viewpoints (zoom-in/zoom-out operations).  
While using 3D smoothing filters like those proposed in Mip-Splatting might seem like a possible solution, they fail to account for local motion and also exhibit aliasing. 
To address this, we propose a motion-aware 3D smoothing filter specifically designed for 4D representations. Our approach adapts the filter strength based on local motion information, effectively mitigating aliasing without compromising rendering quality. This is achieved by estimating the joint density function of time and focal-to-depth ratio using a non-parametric estimation method. During inference, we sample from this joint distribution to determine the appropriate smoothing filter. This flexible strategy can be integrated with various 4D representations.  
Our evaluations on standard datasets demonstrate superior performance compared to state-of-the-art methods. 
\vspace{-5mm}
\end{abstract}
    
\section{Introduction}
\label{sec:intro}

\begin{figure}
    \centering
    \includegraphics[width=\linewidth]{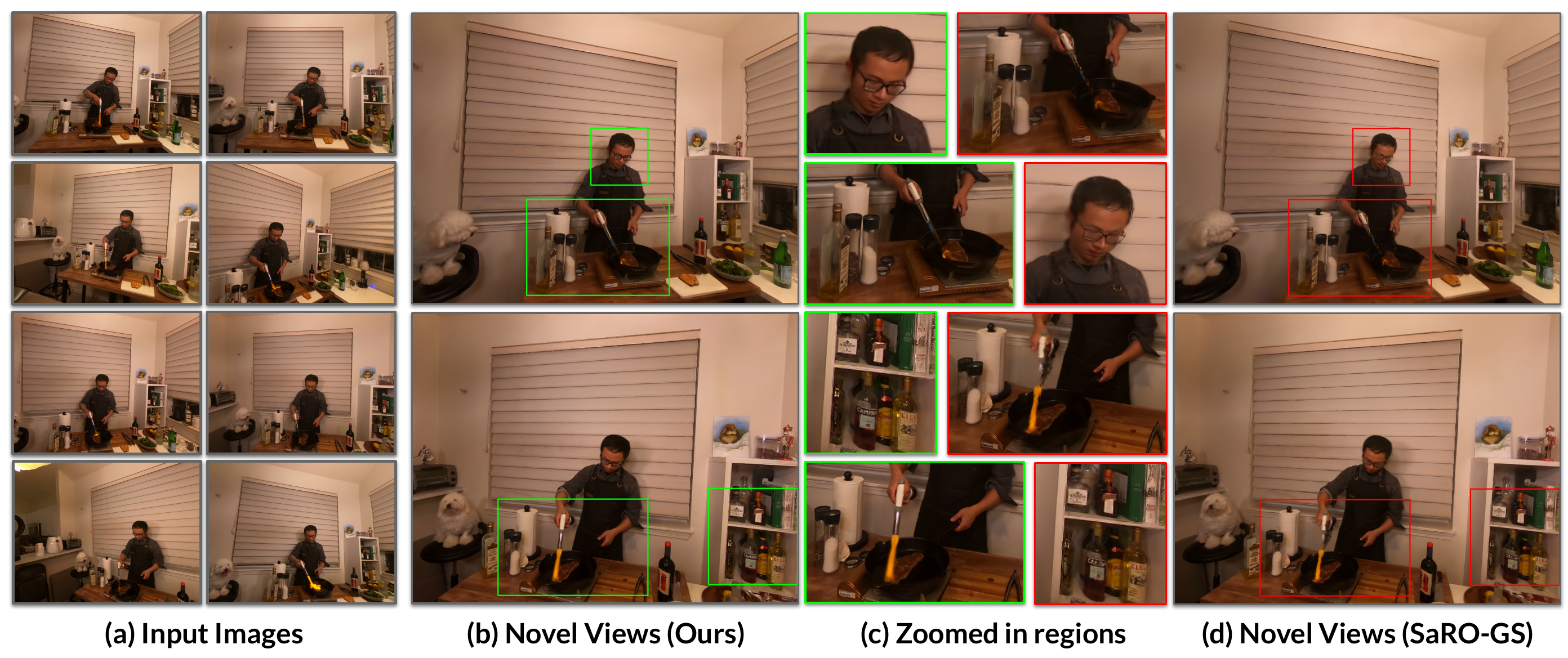}
    \caption{We compare novel view synthesis results from our method (b) with SARO-GS (d), focusing on both dynamic regions (e.g., hand movements, cooking actions) and static regions (e.g., bottles, background). The zoomed-in regions (c) highlight that our method produces sharper details and fewer artifacts, demonstrating superior synthesis quality in motion and static areas.}
    \label{fig:teaser}
\end{figure}
Learning the radiance field~\cite{nerf} of a scene from multi-view RGB images enables numerous applications, including novel-view synthesis (NVS)~\cite{barron2021mip,barron2022mip,muller2022instant,dhiman2023strata,3dgs}, 3D reconstruction~\cite{sugar,gaussianfrosting}, mixed reality~\cite{henriques2024foveated}, and robotics~\cite{soti20246,wang20246}. 
Generating novel views of dynamic scenes from multi-view camera observations holds immense potential for creating immersive experiences from everyday video recordings~\cite{gao2021dynamic}. However, it is very challenging.
Traditional NVS methods relied on implicit~\cite{nerf, barron2021mip,barron2022mip} or grid-based~\cite{fridovich2022plenoxels,muller2022instant,barron2023zip} representations, often employing ray-tracing and volumetric rendering. More recently, explicit primitive-based representations have become popular. In particular, 3D Gaussian Splatting (3DGS)~\cite{3dgs} utilizes a differentiable rendering pipeline based on Gaussian splat rasterization, achieving real-time rendering of 1080p images on commercial GPUs.

To model dynamic scenes, extensions of NeRF and 3DGS employ the concept of deformation fields and canonical representations~\cite{guo2023forward,liu2022devrf,hypernerf,pumarola2021d,yang2024deformable, guo2024motion, liang2023gaufre}. Alternative methods decompose the 4D scene into 3D projections of spatial-temporal and spatial-only 3D representations~\cite{cao2023hexplane,fridovich2023k,lin2023high,shao2023tensor4d,wu20244d,sarogs}, subsequently combining these reduced features to model the dynamic regions. Other methods extend the\begin{wrapfigure}{r}{0.55\textwidth}
    \vspace{-10pt}
    \centering
    \includegraphics[width=\linewidth]{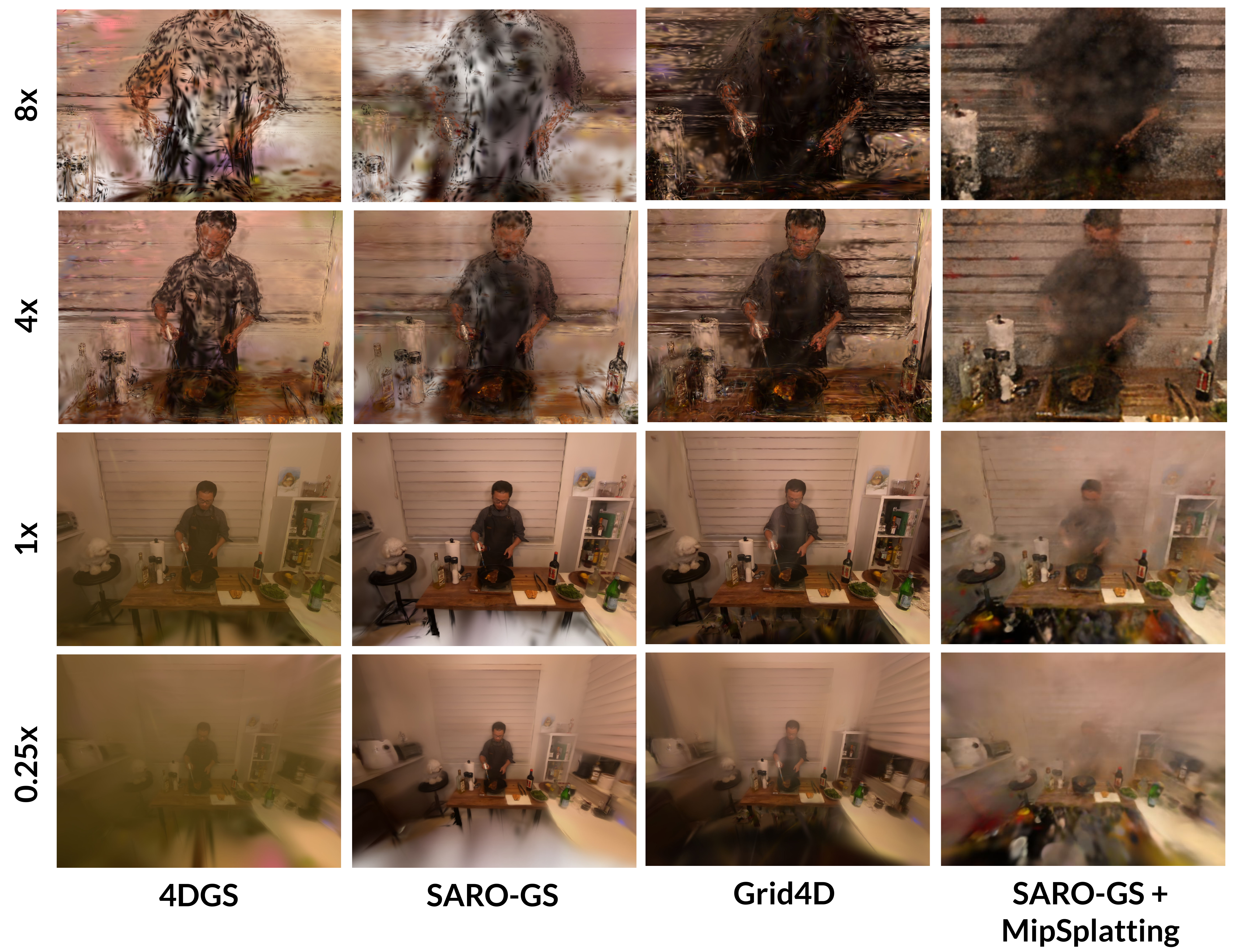}
    \caption{
    We trained all models on training views with a resolution of $1352\times1014$ and rendered outputs by adjusting the focal length. Existing 4D representations exhibit significant artifacts during zoom-in and zoom-out operations. Applying static anti-aliasing filters, such as Mip-Splatting~\cite{mipsplatting}, is also ineffective as motion in the scene leads to incorrect filter estimation and persistent artifacts.} 
    \label{fig:aliasing_problem}
    \vspace{-7mm}
\end{wrapfigure}
 3DGS representation to 4D Gaussians~\cite{das2023neural, katsumata2024compact, duan20244d}, directly optimizing them.  
However, in Fig.~\ref{fig:aliasing_problem} all these representations exhibit limitations, notably aliasing  artifacts when rendered from divergent viewpoints, which hinders accurate representation of dynamic scenes.

Recent work, Mip-Splatting~\cite{mipsplatting}, addresses the challenge of divergence in the novel-view viewpoints from training viewpoints (zoom-in and zoom-out operations) for static scenes. Specifically, it determines the maximum sampling interval for each Gaussian primitive and restricts the maximum frequency of the 3D representation by applying a low-pass filter to satisfy the Nyquist criterion~\cite{nyquist}. While demonstrating impressive results for static scenes, this methodology exhibits limitations for dynamic scene representations (See Fig.~\ref{fig:aliasing_problem}). A primary reason for this limitation is that the band-limited filter lacks motion awareness, failing to account for temporal changes in the scene. Consequently, we propose a motion-aware strategy to determine the maximum sampling interval, thereby enabling accurate rendering for dynamic scenes.

To mitigate aliasing artifacts observed in dynamic scenes, we first analyze the temporal distribution of the sampling interval to derive a motion-aware smoothing filter. 
We found that directly estimating the sampling interval at each timestamp led to noisy estimates due to temporal abruptness. To address this, we propose learning the joint distribution of discrete time and the focal-length-to-depth ratio for each Gaussian primitive. We employ a non-parametric estimation technique: the Parzen Window kernel density estimation method, to learn this joint distribution from accumulated data points. This also makes the estimates smoother compared to the trivial solution. During inference, we sample the most probable value from this learned distribution. 
Overall, this strategy makes the sampling interval estimation motion-aware and enables the design of a motion-aware smoothing filter for these 4D representations.
Our main contributions are as follows:
\begin{itemize}
    \item We introduce a motion-aware smoothing filter that resolves the aliasing artifacts in the 4D representations.
    \item We demonstrate the efficacy of our method in challenging real-world datasets~\cite{li2022neural,hypernerf} and a synthetic dataset~\cite{pumarola2021d}.
    \item Our modifications are invariant of the 4D representation and can be plugged in with other representations as well.
\end{itemize}

\section{Related Work}
\label{sec:related_work}

\noindent
\textbf{Novel View Synthesis.}
With the advancement in neural rendering steered by Neural Radiance Fields (NeRF)~\cite{nerf}, the task of novel-view synthesis has gained a lot of popularity. NeRF utilizes volumetric rendering and optimizes a multi-layer perception (MLP) to learn the radiance field of a scene. This approach is computationally expensive. To accelerate training and rendering speed subsequent works makes use of hash grids~\cite{muller2022instant}, voxel grids~\cite{sun2022direct,fridovich2022plenoxels}, point-based methods~\cite{xu2022point} and depth supervision~\cite{deng2022depth}. Further, this representation is extended to solve different tasks such as large-scale scenes~\cite{rematas2022urban,xiangli2022bungeenerf}, text-to-3D generation~\cite{tang2023dreamgaussian, wang2024prolificdreamer}, hierarchical scenes~\cite{dhiman2023strata} and editing tasks~\cite{huang2022stylizednerf,zhang2022arf,dhiman2024chroma}.

\begin{figure}[!t]
    \centering
    \includegraphics[width=\linewidth]{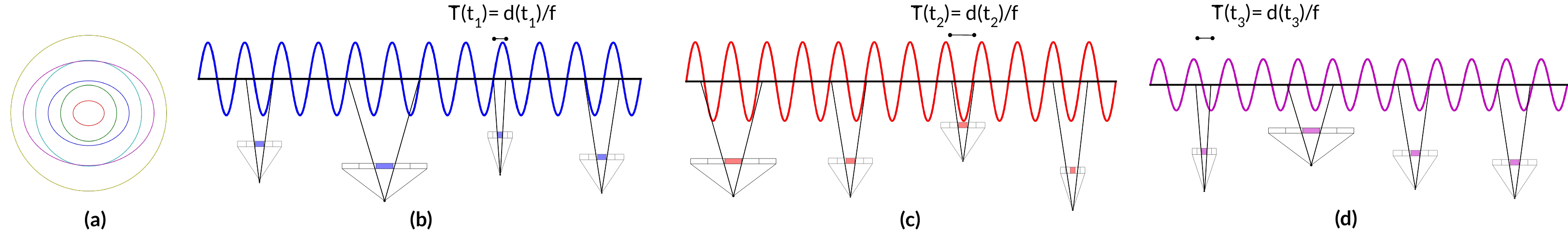}
    \caption{Consider a 4D signal (represented as a 2D signal in (a)). Its different projections at timestamps  $t_1\,,t_2\,\&\,t_3$, form 3D signals (represented as 1D signals in (b), (c), and (d), respectively). These continuous 3D signals are modeled by 3D Gaussian primitives. When observed by multiple cameras at each timestamp, these primitives are band-limited by different sampling intervals. Hence, it is essential to design a motion-aware low-pass filter for dynamic scenes.} 
    \label{fig:sampling_limits}
    \vspace{-1.5em}
\end{figure}

3DGS~\cite{3dgs} models scenes as a mixture of 3D Gaussians that are projected as 2D splats and rasterized using alpha blending to get the projected image. Compared to NeRF based approaches, 3DGS enables high-quality rendering in real-time. Subsequent works have improved over this representation by introducing a hierarchical structure~\cite{lu2024scaffold}, generalizable exponential functions~\cite{hamdi2024ges}. Further, other works have focused on improving rendering quality~\cite{huang20242d,mipsplatting}, accelerated rendering~\cite{girish2024eagles,fan2023lightgaussian}, memory optimization~\cite{papantonakis2024reducing,navaneet2024compgs}, better initialization~\cite{jung2024relaxing,kheradmand20243d,paliwal2024coherentgs} and overall training time~\cite{girish2024eagles, hollein20243dgs, mallick2024taming,dhiman2026turbo}.   

\noindent
\textbf{Dynamic Scenes.}
To model dynamic scenes, NeRF based methods~\cite{guo2023forward,liu2022devrf,hypernerf,pumarola2021d} model a scene as a canonical scene and a deformed scene and then use deformation field network to model the relationship between static canonical space and the deformed points. In contrast,~\cite{cao2023hexplane,fridovich2023k,lin2023high,shao2023tensor4d, wang2024masked} decomposes the 4D space into set of planes or hash-grids to effectively model the spatio-temporal relations. However, these approaches require dense sampling and suffer from same issue as the static scene counterparts such as low rendering speed. 

Building on 3DGS, recent methods extend to dynamic scenes. Dynamic 3D Gaussians~\cite{luiten2024dynamic} model a dynamic scene as a collection of moving and rotating Gaussians frame by frame. 4DGS~\cite{wu20244d} utilizes Hex-planes to model the changes in
Gaussian primitives over time. Other methods~\cite{yang2024deformable, guo2024motion, liang2023gaufre} utilize a time-conditioned deformation network to wrap a canonical set of Gaussians into each timeframe. These methods fail to handle scenarios with significant motion and often suffer from artifacts such as object appearances/disappearances. Further works~\cite{das2023neural, katsumata2024compact, duan20244d} extend 3D Gaussians across space-time, i.e. means and covariances of Gaussian primitives are in 4D. Other works~\cite{stearns2024dynamic,das2023neural,katsumata2024compact} attempt the challenging task of novel-view synthesis of casual monocular videos for dynamic scenes. In comparison, our work addresses the aliasing artifacts in these representations. 

\noindent
\textbf{Aliasing in Radiance Field Methods.} 
Earlier NeRF-based~\cite{barron2021mip,barron2022mip,barron2023zip,hu2023tri,zhuang2023anti} approaches used pre-filtering to mitigate aliasing, but these methods do not extend to 3DGS representations. EWA splatting~\cite{zwicker2001ewa} filters 2D Gaussians in screen space, but this rendering-time filter is prone to the choice of filter size.
Conversely, Mip-Splatting~\cite{mipsplatting} designs a 3D filter during optimization, allowing alias-free rendering at different scales, but this fails for dynamic scenes. In contrast, we propose applying a 3D band-limited filter based on the Gaussian primitive's motion.

\vspace{-3mm}
\section{Preliminaries}
\label{sec:preliminary}

In this section, we provide a brief overview of 3D Gaussian Splatting (3DGS)~\cite{3dgs} and a recent dynamic representation SARO-GS~\cite{sarogs}. We then discuss Mip-Splatting~\cite{mipsplatting}, an anti-aliasing 3D filter for static scenes.

\subsection{3D Gaussian Splatting}

3DGS~\cite{3dgs} represents scene geometry using a collection of 3D Gaussian primitives and each primitive is defined as:
\begin{equation}
    G(x)=e^{-\frac{1}{2}(x-\mu)^T \Sigma^{-1}(x-\mu)}
\end{equation}
where $\mu$ is the center position and $\Sigma$ is an anisotropic covariance matrix. A combination of rotational $R$ and scaling $S$ transformations, represented as $RSS^TR^T$, constructs the covariance matrix, guaranteeing that it is positive semi-definite. Each Gaussian primitive has an opacity value $\sigma$, and spherical harmonic coefficients to represent a view-dependent color. During rendering, Gaussians are projected as 2D splats, sorted by depth, and combined with $\alpha$ blending using a tile-based rasterizer. The resulting pixel color $C$, is determined by:
\begin{equation}
    C\left(x^{\prime}\right)=\sum_{i}c_i \sigma_i \prod_{j=1}^{i-1}\left(1-\sigma_j\right), \quad
    \sigma_i=\alpha_i G_i^{\prime}\left(x^{\prime}\right)
    \label{eq:rasterization}
\end{equation}
where $x^{\prime}$ is the queried pixel, $c_i$ is the color of the $i$-th Gaussian, $\alpha_i$ is the learned opacity for the $i$-th Gaussian, and $G_i^{\prime}\left(x^{\prime}\right)$ is the 2D projection of the Gaussian at pixel $x^{\prime}$.

\subsection{Dynamic Representations using 3DGS}
\label{sec:pre_hexplane}
We choose SARO-GS~\cite{sarogs} as the base 4D representation. SARO-GS models a dynamic scene using a set of 4D Gaussian primitives, $\mathcal{G}^{4D}$, and a Scale-aware Residual Field, $\mathcal{M}$. Each 4D Gaussian $\mathcal{G}^{4D}$ is parameterized by a 4D location $\mu^{4D}=(\mu,\tau)$ and other attributes covariance $\Sigma^{4D}$, color $c^{4D}$ and opacity $\sigma^{4D}$. Here, $\mathcal{M}$ is a lightweight MLP that encodes the spatio-temporal variation of the scene: for every primitive it predicts a time-dependent residual to its attributes. Concretely, the 4D Gaussian primitives are first projected into 3D space at the initial time $t_0$. Then, $\mathcal{M}$ is queried at a given time-stamp $t$ to obtain residual features, which are added to the projected 3D Gaussian primitives before rasterization using Eq.~\ref{eq:rasterization}. The final loss is given as follows:
\begin{equation}
\mathcal{L} = (1-\lambda_1)\mathcal{L}_1 + \lambda_1\mathcal{L}_{D-SSIM} +\lambda_2 \mathcal{L}_{SR}
    \label{eq:loss}
\end{equation}
where $\mathcal{L}_1$ and $\mathcal{L}_{D\text{-}SSIM}$ are the standard photometric losses used in 3DGS~\cite{3dgs}, computed between the ground-truth and the rasterized image, and $\mathcal{L}_{SR}$ is the regularization term on the scaling residuals introduced by SARO-GS~\cite{sarogs}.

\begin{algorithm}[!t]
\small
\caption{Motion-Aware Sampling Rate Estimation (per Gaussian primitive)}
\label{algo:motion-aware}
\begin{algorithmic}[1]
\REQUIRE Observations $\{(t_j, x_j)\}_{j=1}^{J}$, $x_j=f_j/d_j$ (empty slots flagged by $t_j<0$); bandwidths $h_x,h_t$; Weibull shape $k$, scale $\lambda$; grid sizes $D,T$
\ENSURE Sampling rate $\hat{\nu}(t)$ for a query time $t$
\STATE Kernels: $K_w(u)=\frac{k}{\lambda}(\frac{u}{\lambda})^{k-1}e^{-(u/\lambda)^{k}}\mathds{1}[u\ge0]$, \ $K_g(u)=\frac{1}{\sqrt{2\pi}}e^{-u^2/2}$
\STATE $v_j \gets \mathds{1}[t_j \ge 0]$ \COMMENT{mask empty observation slots}
\STATE $x_{\max} \gets \max_{j} v_j\,x_j$; \ $\bar{x}_j \gets x_j / x_{\max}$ \COMMENT{normalize $f/d$ to $[0,1]$ per primitive}
\STATE $g^x \gets \textsc{Linspace}(0,1,D)$; \ $g^t \gets \textsc{Linspace}(0,1,T)$ \COMMENT{query grids}
\FOR{$d=1$ to $D$}
    \FOR{$s=1$ to $T$}
        \STATE $\rho[d,s] \gets \frac{1}{2 h_x h_t}\sum_{j=1}^{J} v_j\, K_w\!\left(\frac{g^x_d-\bar{x}_j}{h_x}\right) K_g\!\left(\frac{g^t_s-t_j}{h_t}\right)$ \COMMENT{joint KDE}
    \ENDFOR
\ENDFOR
\STATE \textit{Inference at a query time} $t$:
\STATE \quad $s^\star \gets \arg\min_{s} |g^t_s - t|$ \COMMENT{nearest time grid point}
\STATE \quad $d^\star \gets \arg\max_{d} \rho[d, s^\star]$ \COMMENT{mode of the conditional density $\rho[\cdot,s^\star]$}
\STATE \quad $\hat{\nu}(t) \gets x_{\max}\cdot g^x_{d^\star}$ \COMMENT{rescale to $f/d$ units}
\end{algorithmic}
\end{algorithm}

\subsection{Anti-aliasing filter for 3DGS}
\label{sec:pre_mipsplat}
Mip-Splatting~\cite{mipsplatting} builds upon the principle that image resolution, camera focal length, and scene depth define the effective sampling rate for 3D scene reconstruction from multi-view images. Their major insight is that primitive smaller than twice the world-space sampling interval $\hat{T}$ can introduce aliasing, where $\hat{T} = \frac{d}{f}$, where $d$ is distance from the camera and $f$ is the focal length of the camera. To mitigate this, they approximate the depth-dependent sampling rate for each primitive based on its center and camera visibility and is given by following: 
\begin{equation}
\hat{\nu}_k = \text{max}\left(\left\{ \mathds{1}_n(\bp_k) \cdot \frac{f_n}{d_n}\right\}^{N}_{n=1}\right)
 \label{eqn:sample_interval}
\end{equation}
where $N$ is the total number of images, $\mathds{1}_n(\bp)$ is an indicator function that assesses the visibility of a primitive. Then, they apply a Gaussian low-pass filter $\cG_{\text{low}}$ to a 3D Gaussian primitive $\cG_k$ before projecting it onto screen space:
\begin{align}
\cG_k(\bx)_{\text{reg}} = (\cG_k \otimes \cG_{\text{low}})(\bx)
\label{eqn:filter_3d_simple}
\end{align}
Result of convolving two Gaussians is a Gaussian and is given by:
\begin{equation}
\cG_k(\bx)_\text{reg} = \sqrt{\frac{|\bSigma_k|}{|\bSigma_k + \frac{s}{\hat{\nu}_k} \cdot \bI|}} \, \, e^{-\frac{1}{2} (\bx-\mu)^T \, (\bSigma_k + \frac{s}{\hat{\nu}_k} \cdot \bI)^{-1} \, (\bx-\mu)}
\label{eqn:filter_3d_full}
\end{equation}
where $s$ controls the size of the filter.  Note that $\frac{s}{\hat{\nu}_k}$ is different for each Gaussian primitive and remains constant post-training.

\section{Proposed Method}
\label{sec:method}

\noindent
\textbf{Aliasing in Dynamic Representations }
Fig.~\ref{fig:aliasing_problem} demonstrates the aliasing artifacts present in these 4D representations. After training 4D representations on input views, we perform zoom-in and zoom-out operations. During zoom-in, 4DGS~\cite{wu20244d}, SARO-GS~\cite{sarogs}, and Grid4D~\cite{grid4d} suffer from disappearing primitives in both dynamic and static regions. During zoom-out, these methods produce noticeable aliasing. For example, in Grid4D, the human merges with the wall, and SARO-GS misses fine floor details. These results indicate that aliasing occurs in 4D representations when novel-view camera viewpoints deviate significantly from the training data.

\noindent
\textbf{Limitations of Static Anti-Aliasing Filters}
Aliasing artifacts pose a significant challenge in reconstructing dynamic scenes with dynamic representations, \begin{wrapfigure}{r}{0.5\textwidth}
    \centering
   \includegraphics[width=\linewidth]{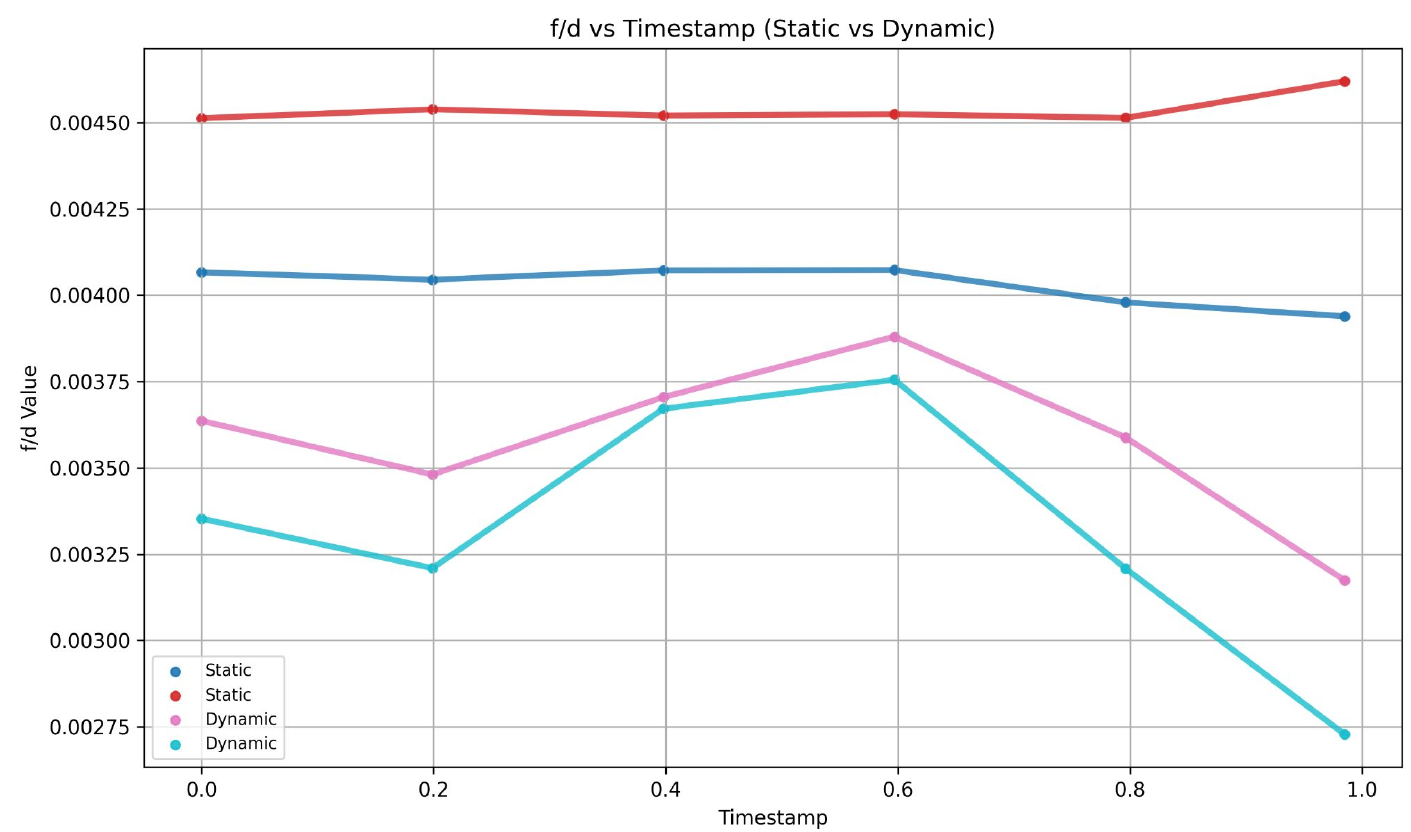}
    \caption{Analysis of the sampling rate $f/d$ over time for the ``cook-spinach'' scene from the Plenoptic Video dataset~\cite{li2022neural}. Static primitives maintain a nearly constant $f/d$, whereas dynamic primitives vary substantially over time, as motion changes their depth. We detail how static and dynamic primitives are identified in the supplementary material.}
    \label{fig:dist_d/f}
    \vspace{-5mm}
\end{wrapfigure}
particularly in areas of motion (Fig.~\ref{fig:aliasing_problem}). While Mip-Splatting~\cite{mipsplatting} effectively addresses aliasing in static scenes by introducing a low-pass filter, it proves ineffective for dynamic scenarios, as evidenced by the results for ``SARO-GS + Mip-Splatting'' in Fig.~\ref{fig:aliasing_problem}. Therefore, we propose a motion-aware smoothing filter tailored for dynamic scene reconstruction.

\noindent
\textbf{Analysis of Static and Dynamic Gaussian primitives.}
Analysis of the sampling rate $(f/d)$ distribution 
over time reveals key insights for dynamic scenes. As presented in Fig.~\ref{fig:dist_d/f} for the ``cook-spinach'' scene from the Plenoptic Video dataset~\cite{li2022neural}, static regions exhibit a consistently stable distribution, as expected. In contrast, dynamic regions show significant temporal variation, demonstrating that a single, static smoothing filter is inadequate and results in artifacts (Fig.~\ref{fig:aliasing_problem}). This observation directly motivates the design of our motion-aware smoothing filter.

\begin{figure}[!t]
  \centering
  \includegraphics[width=\linewidth]{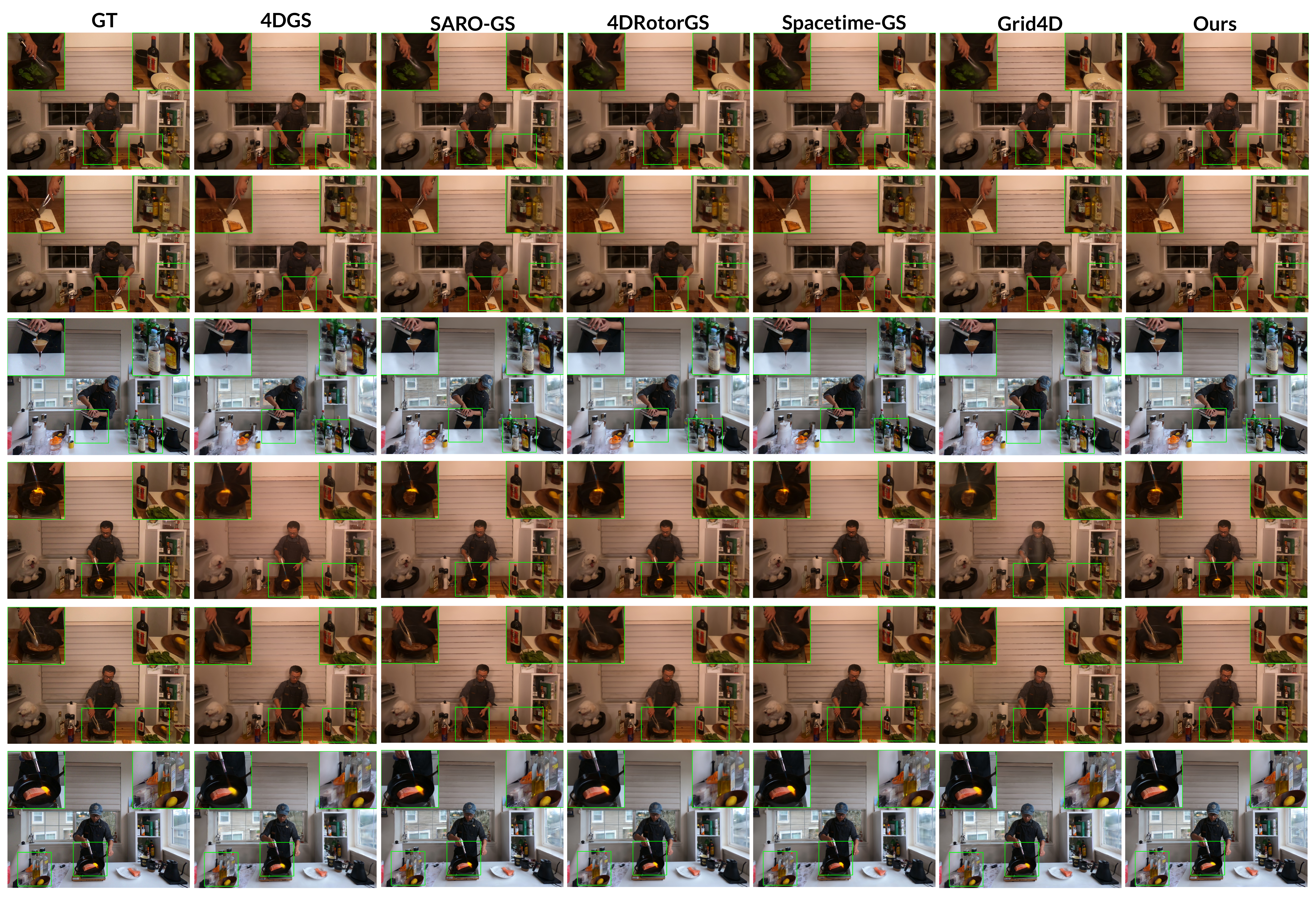}
   \caption{\textbf{Single-Scale Training and Multi-Scale Evaluation on Plenoptic Video~\cite{li2022neural} dataset.} 
  All baseline methods and our approach were trained at a resolution downscaled by a factor of $4$, while novel views are evaluated at the original. We provide zoomed-in regions highlighted by green boxes. Notice that the fine details of the martini (Row 3) being poured into the glass are not preserved in any baseline method, whereas our method successfully retains this detail.
  }
  \label{fig:qual_plenoptic}
\end{figure}

\noindent
\textbf{Necessity of Time-Varying Sampling Rates.}
Fig.~\ref{fig:sampling_limits} illustrates the challenge of estimating sampling intervals for dynamic scenes. Unlike static scenes, dynamic scenes are continuous 4D signals. While Mip-Splatting can provide an overall estimate of the maximal sampling rate for this 4D signal, a single value is insufficient. Since each timestamp $t$ yields a unique 3D projection of the 4D signal, it is necessary to determine a separate maximal sampling rate for each projection. Therefore, we propose a novel method to estimate the maximal sampling interval for a timestamp. We emphasize that here ``sampling rate'' denotes the \emph{spatial} sampling requirement $f/d$ of a primitive at a given timestamp. Although the video timestamps are fixed, a primitive's depth changes with motion, which makes its $f/d$ time-varying. Our method thus computes a separate maximal sampling rate for each spatio-temporal projection of the 4D signal, allowing the anti-aliasing filter to adapt per primitive and per timestep while satisfying the Nyquist criterion.


\noindent
\textbf{Motion-Aware Smoothing Filter}
We assume that $f/d$ is sampled from a distribution $\cX$ and $t$ is sampled from a distribution $\cT$. We do not know the underlying distribution of these variables and want to estimate the joint density function $\cP(\cX,\cT)$. Hence, we utilize the Parzen window density estimation method to estimate the density function $\cP(\cX,\cT)$. Parzen window methods yield smooth and continuous density estimates, in contrast to histograms which are discontinuous and sensitive to bin size. We choose following kernel functions:
\setlength{\abovedisplayskip}{0.5\abovedisplayskip}
\setlength{\belowdisplayskip}{0.5\belowdisplayskip}
\begin{equation}
    \cX \sim Weibull( k,\lambda) \,\,\,\,\,\,\, \cT \sim \mathcal{N}(\mu_T, \sigma_T^2)
\end{equation}
We assume a Weibull distribution because, for 
$k > 1$, it becomes biased toward maximum values. This aligns with Eq.~\ref{eqn:sample_interval}, where the sampling rate for each primitive is defined as the maximum of all entries.

\begin{table}[!t]
    \renewcommand{\tabcolsep}{1pt}
    \centering
    \resizebox{0.95\linewidth}{!}{
    \begin{tabular}{@{}l@{\,\,}|cccc|cccc|cccc}
    \toprule
    & \multicolumn{4}{c|}{PSNR $\uparrow$} & \multicolumn{4}{c|}{SSIM $\uparrow$} & \multicolumn{4}{c}{LPIPS $\downarrow$}  \\
    & $1 \times$ Res. & $2 \times$ Res. & $4 \times$ Res.  & Avg. & $1 \times$ Res. & $2 \times$ Res. & $4 \times$ Res. &  Avg. & $1 \times$ Res. & $2 \times$ Res. & $4 \times$ Res. &  Avg.  \\ \hline
    SARO-GS~\cite{sarogs}          &                        
                              \cellcolor{tablered}32.75 &   
                              \cellcolor{tableyellow}29.40 &
                              \cellcolor{tableyellow}27.40 &
                              \cellcolor{tableyellow}29.85 &
                              \cellcolor{tableorange}0.956 &   
                              0.913 &
                              \cellcolor{tableyellow}0.884 &
                              \cellcolor{tableyellow}0.918 &
                              \cellcolor{tablered}0.086 &   
                              \cellcolor{tableorange}0.167 &
                              0.234 &
                              \cellcolor{tableorange}0.160 \\
Spacetime-GS~\cite{spacetimegaussians}                            & 
                              \cellcolor{tableyellow}32.50 &   
                              29.20 &
                              27.09 &
                              29.60 &
                              \cellcolor{tablered}0.957 &   
                              \cellcolor{tableyellow}0.918 &
                              \cellcolor{tableorange}0.892 &
                              \cellcolor{tableorange}0.922 &
                              \cellcolor{tableorange}0.087 &   
                              \cellcolor{tableyellow}0.169 &
                              0.232 &
                              0.163 \\
4DGS~\cite{wu20244d}                                &                       
                              31.46 &   
                              28.30 &
                              26.36 &
                              28.71 &
                              0.946 &   
                              0.901 &
                              0.881 &
                              0.909 &
                              0.095 &   
                              0.177 &
                              \cellcolor{tableyellow}0.230 &
                              0.167 \\
4DRotorGS~\cite{duan20244d}                          &                       
                              32.29 &   
                              \cellcolor{tableorange}29.60 &
                              \cellcolor{tableorange}29.10 &
                              \cellcolor{tableorange}30.33 &
                              0.951 &   
                              \cellcolor{tableorange}0.920 &
                              0.877 &
                              0.916 &
                              0.092 &   
                              0.187 &
                              \cellcolor{tablered}0.212 &
                              \cellcolor{tableyellow}0.162 \\
Grid4D~\cite{grid4d}   &                       
                              31.03 &   
                              28.13 &
                              26.64 &
                              28.60 &
                              0.799 &   
                              0.773 &
                              0.765 &
                              0.779 &
                              0.095 &   
                              0.177 &
                              0.234 &
                              0.169 \\
Ours   &                       
                              \cellcolor{tableorange}32.52 &   
                              \cellcolor{tablered}30.75 &
                              \cellcolor{tablered}29.79 &
                              \cellcolor{tablered}31.01 &
                              \cellcolor{tableyellow}0.952 &   
                              \cellcolor{tablered}0.925 &
                              \cellcolor{tablered}0.903 &
                              \cellcolor{tablered}0.926 &
                              \cellcolor{tableyellow}0.089 &   
                              \cellcolor{tablered}0.162 &
                              \cellcolor{tableorange}0.216 &
                              \cellcolor{tablered}0.155 
    \end{tabular}
    }
    \caption{
    \textbf{Single-Scale Training and Multi-Scale Evaluation on the Plenoptic Video Dataset~\cite{li2022neural}.} All methods are trained at the smallest scale ($1 \times$) that is $676$x$507$ and evaluated across multiple scales ($1 \times$, $2 \times$, and $4 \times$), where increasing resolution simulates zoom-in effects. While our approach performs comparably to existing methods at the training resolution ($1 \times$), it demonstrates superior fidelity across all other scales, significantly outperforming prior work.}   \label{tab:avg_plenoptic_single_train_multi_test}
\end{table}

\begin{table}[!t]
    \renewcommand{\tabcolsep}{1pt}
    \centering
    \resizebox{\linewidth}{!}{
   \begin{tabular}{@{}l@{\,\,}|cccc|cccc|cccc}
   \toprule
    & \multicolumn{4}{c|}{PSNR $\uparrow$} & \multicolumn{4}{c|}{SSIM $\uparrow$} & \multicolumn{4}{c}{LPIPS $\downarrow$}  \\
    & $1 \times$ Res. & $2 \times$ Res. & $4 \times$ Res.  & Avg. & $1 \times$ Res. & $2 \times$ Res. & $4 \times$ Res. &  Avg. & $1 \times$ Res. & $2 \times$ Res. & $4 \times$ Res. &  Avg.  \\ \hline
    Deformable3DGS~\cite{yang2024deformable} & 
                              \cellcolor{tableorange}39.57 &   
                              \cellcolor{tableorange}30.09 &
                              \cellcolor{tableyellow}27.30 &
                              \cellcolor{tableyellow}32.32 &
                              \cellcolor{tableorange}0.989 &   
                              \cellcolor{tableyellow}0.957 &
                              \cellcolor{tableyellow}0.934 &
                              \cellcolor{tableorange}0.960 &
                              \cellcolor{tableorange}0.008 &   
                              \cellcolor{tableorange}0.034 &
                              \cellcolor{tableyellow}0.061 &
                              \cellcolor{tablered}0.034 \\
SARO-GS~\cite{sarogs}          &                        
                              36.61 &   
                              29.59 &
                              27.15 &
                              31.12 &
                              0.982 &   
                              0.952 &
                              0.931 &
                              \cellcolor{tableyellow}0.955 &
                              0.015 &   
                              0.040 &
                              0.064 &
                              \cellcolor{tableyellow}0.039 \\

4DGS~\cite{wu20244d}                                &                       
                              35.65 &   
                              28.03 &
                              25.58 &
                              29.75 &
                              \cellcolor{tableyellow}0.987 &   
                              \cellcolor{tableorange}0.958 &
                              \cellcolor{tableorange}0.937 &
                              \cellcolor{tableorange}0.960 &
                              \cellcolor{tableyellow}0.012 &   
                              \cellcolor{tableyellow}0.035 &
                              \cellcolor{tableorange}0.059 &
                              \cellcolor{tableorange}0.035 \\
4DRotorGS~\cite{duan20244d}                          &                       
                              32.17 &   
                              26.77 &
                              24.84 &
                              27.92 &
                              0.965 &   
                              0.931 &
                              0.919 &
                              0.938 &
                              0.024 &   
                              0.050 &
                              0.075 &
                              0.050 \\
Grid4D~\cite{grid4d}   &                       
                              \cellcolor{tablered}40.30 &   
                              \cellcolor{tableyellow}30.07 &
                              \cellcolor{tableorange}28.54 &
                              \cellcolor{tableorange}32.97 &
                              \cellcolor{tablered}0.990 &   
                              0.956 &
                              0.933 &
                              \cellcolor{tableorange}0.960 &
                              \cellcolor{tablered}0.007 &   
                              \cellcolor{tableyellow}0.035 &
                              0.062 &
                              \cellcolor{tableorange}0.035 \\
GauFRe~\cite{gaufre}   &                       
                              34.05 &   
                              27.59 &
                              25.12 &
                              28.92 &
                              0.981 &   
                              0.951 &
                              0.928 &
                             0.953 &
                             0.025 &   
                              0.058 &
                              0.079 &
                              0.054 \\
Ours   &                       
                              \cellcolor{tableyellow}37.10 &   
                              \cellcolor{tablered}34.05 &
                              \cellcolor{tablered}32.36 &
                              \cellcolor{tablered}34.50 &
                              0.982 &   
                              \cellcolor{tablered}0.972 &
                              \cellcolor{tablered}0.960 &
                              \cellcolor{tablered}0.971 &
                              0.017 &   
                              \cellcolor{tablered}0.033 &
                              \cellcolor{tablered}0.054 &
                              \cellcolor{tableorange}0.035 

    \end{tabular}
    }
    \caption{
    \textbf{Single-Scale Training and Multi-Scale Evaluation on the Synthetic Dataset~\cite{pumarola2021d}.} All methods are trained on the smallest scale ($1 \times$) that is $200$x$200$ and evaluated across multiple scales ($1 \times$, $2 \times$, and $4 \times$), where increasing resolution simulates zoom-in effects. Our approach outperforms baselines across all other scales. }    \label{tab:avg_synthetic_single_train_multi_test}
    \vspace{-1em}
\end{table}

With these assumptions, we now describe our method. First, for each Gaussian primitive $k$ we create data  $\left\{ (t_i,\frac{f_i}{d_i}) \right\}_{i=1}^{n}$ by projecting primitives at every time stamp for all the training views. To make the estimate invariant to the absolute scale of each primitive's sampling rate, we normalize its $f/d$ values by their per-primitive maximum $x_{\max}$ so that they lie in $[0,1]$, and define the density grids over $[0,1]$; the value recovered at inference is rescaled by $x_{\max}$ to return to $f/d$ units. Then, we estimate the joint density function $\cP_{k}(\cX,\cT)$ for each Gaussian primitive. We describe the algorithm in more detail in Algorithm~\ref{algo:motion-aware}. 
To find the value of $f/d$ given a specific $t$, we determine it maximizing the conditional density $P(\cX|\cT=t)$. Specifically, we extract the estimated density values for a specific $t$; $\cP_{k}(\cX,\cT=t)$ and then find the value corresponding to the highest density.  Finally, this estimated sampling rate is applied to a deformed Gaussian primitive $\cG'_k$ as shown in Eq.~\ref{eqn:filter_3d_simple}, which ensures that the highest frequency component of any Gaussian does not exceed half of its maximal sampling rate.
\section{Experiments}
\label{sec:results}

\noindent
\textbf{Implementation Details}
We use SARO-GS~\cite{sarogs} as the base 4DGS representation. We apply the smoothing filter on the projected 3D Gaussians and train our model for $15$K iterations for all plenoptic scenes and $20$K iterations for synthetic scenes. We use the loss function described in Eq.~\ref{eq:loss}. The adaptive density control happens after $600$ iterations at every $100$th iteration until $5$K iterations for plenoptic scenes and until $15$K iterations for synthetic scenes. We compute the motion-aware smoothing filter after $600$ iterations at every $100$th iteration except the final training iteration. We use the hyperparameter settings similar to SARO-GS. Further, we also use the 2D filter proposed in Mip-Splatting with variance of $0.1$. 

\begin{figure}[!t]
  \centering
  \includegraphics[width=0.90\linewidth]{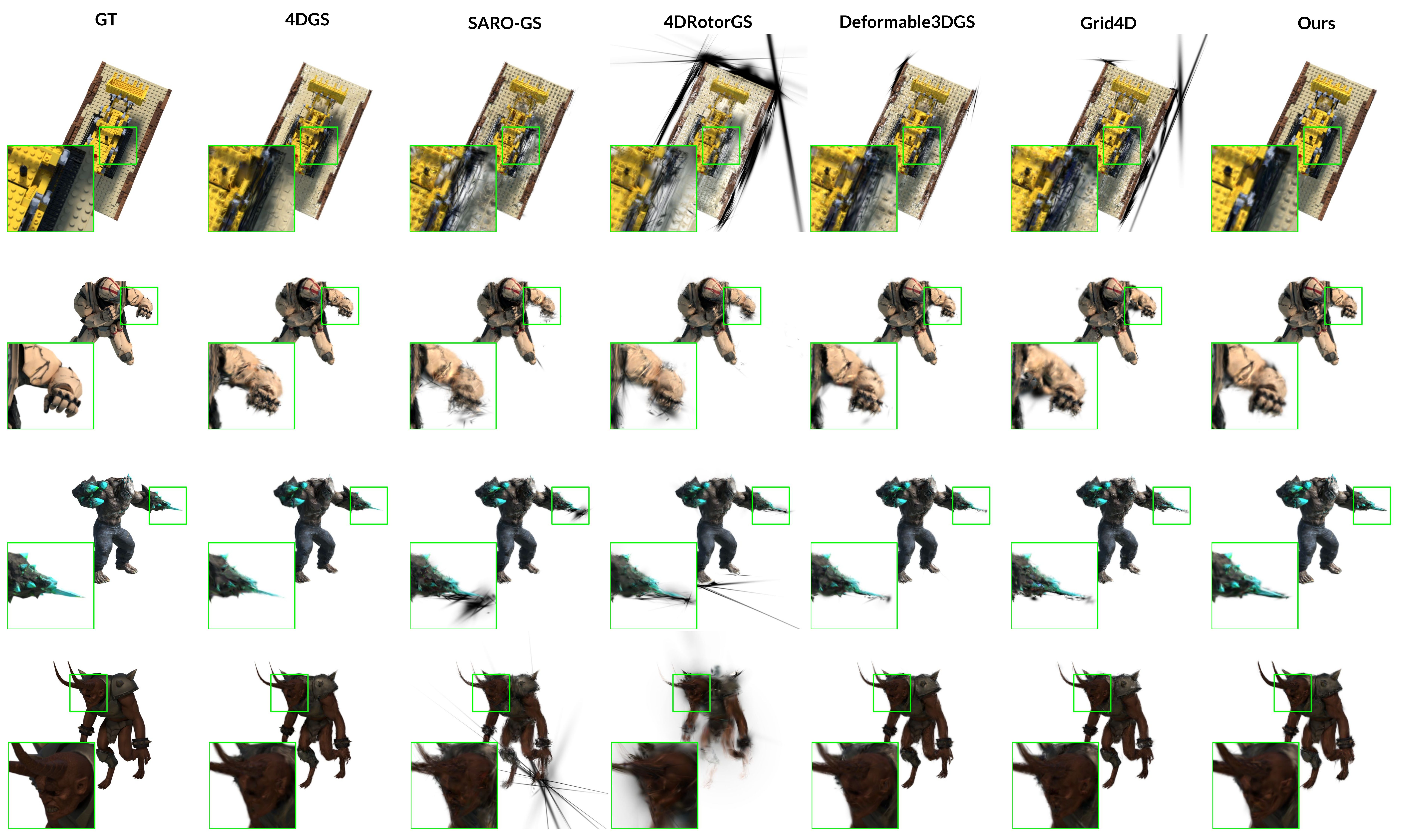}
  \caption{\textbf{Single-Scale Training and Multi-Scale Evaluation on D-NeRF~\cite{pumarola2021d} dataset.} 
  All baseline methods and our method were trained at $200\times200$ image resolution, while novel views are evaluated at $800\times800$. Notice the artifacts in the baselines: the arm in ``mutant'' (Row 3), the arm in ``hook'' (Row 2), and the top part in ``lego'' (Row 1). In contrast, our method renders with high fidelity.
  }
  \label{fig:qual_synthetic}
  \vspace{-4mm}
\end{figure}

\noindent
\textbf{Dataset}
We evaluate our method on six real-world scenes from the Plenoptic Video dataset~\cite{li2022neural}, each captured by 15 to 20 cameras. Additionally, we assess performance on eight synthetic scenes from D-NeRF~\cite{pumarola2021d} dataset. These datasets were utilized for single-scale training and multi-scale testing. We also show evaluation results on the HyperNeRF~\cite{hypernerf} dataset.

\noindent
\textbf{Metrics.} We employ the Peak Signal-to-Noise Ratio (PSNR), Structural Similarity (SSIM)~\cite{wang2004image}, and Learned Perceptual Image Patch Similarity (LPIPS)~\cite{zhang2018unreasonable} to assess the quality of the rendered novel-views. Further, we evaluate the temporal stability of a video based on
the flow warping error between two frames~\cite{lai2018learning}. We provide more details on this temporal stability metrics in the supplementary material.

\noindent
\textbf{Baselines.} We compare our method with the Gaussian splatting-based methods SARO-GS~\cite{sarogs}, 4DGS~\cite{wu20244d}, Spacetime-GS~\cite{spacetimegaussians}, 4DRotorGS~\cite{duan20244d} and Grid4D~\cite{grid4d} and use the official implementations of all baselines.

\subsection{Results and Analysis}

\noindent
\textbf{Single-Scale and Multi-Scale Testing}
To evaluate zoom-in performance, we adopt the methodology from Mip-Splatting. Specifically, we train the model on training-images downsampled by a factor of four. Then, we evaluate the generated novel views  at ($1\times\,,2\times$ and $4\times$) resolution. We train baseline methods using this setting.

\begin{figure}[!t]
    \centering
    \includegraphics[width=\linewidth]{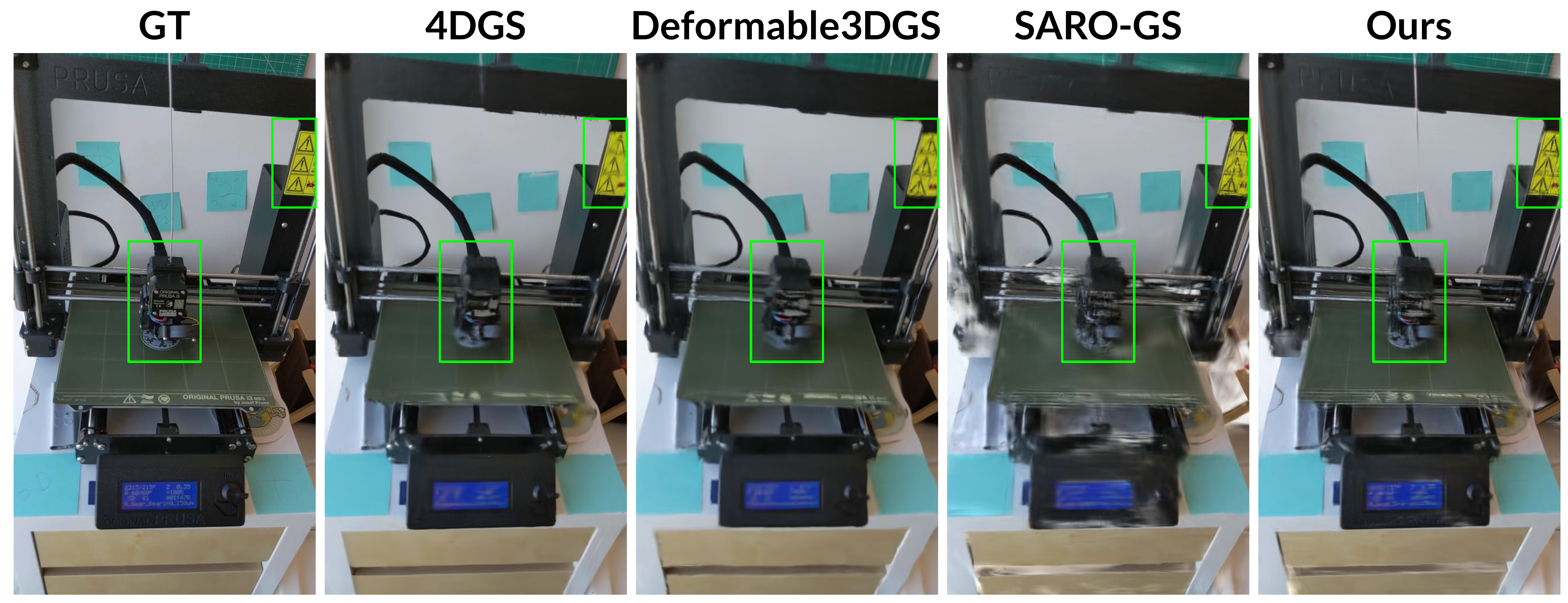}
    \caption{\textbf{Qualitative Results on HyperNeRF~\cite{hypernerf} dataset.} We use the ``Single-Scale Training and Multi-Scale Testing'' setting, where training is performed at a resolution downscaled by a factor of $4$, and novel views are rendered at the original resolution. Notice the blurriness in the ``printer head'' and ``yellow sticker'' (highlighted with a green box) in the baseline methods. In contrast, our method better preserves high-frequency details.}
    \label{fig:hypernerf_small}
    \vspace{-3mm}
\end{figure}

\noindent
\textbf{Evaluation on Plenoptic dataset.} 
Tab.~\ref{tab:avg_plenoptic_single_train_multi_test} shows a multi-scale performance comparison using the training strategy discussed above. At $1\times$ resolution, our method achieves comparable results to the baselines. However, at $2\times$ and $4\times$ resolutions, we significantly outperform the baseline methods. Notably, we achieve a $2.11\,dB$ improvement over SARO-GS, which is significant as we build upon their method. In contrast, 4DGS, Spacetime-GS, and Grid4D exhibit subpar performance at $4\times$ resolution. This trend is also evident in the qualitative results presented in Fig.~\ref{fig:qual_plenoptic}. Observe the blurriness in the baseline methods' output for the ``sear steak'' scene. Our method, on the contrary, preserves high-frequency details when rendered at higher resolutions. This ability to maintain high-frequency content, even when trained on lower-resolution data, is a crucial advantage missing in other 4D representations. Further, our method preserves more high-frequency details than baseline methods on the HyperNeRF dataset in~\cref{fig:hypernerf_small}.

\textbf{Temporal Error Analysis.} We compare the quality of rendered views by evaluating their consistency across consecutive frames in Tab.~\ref{tab:temporal_metrics}. 
This is done by warping a frame to align with the next one and then calculating the error between the warped frame and the actual subsequent frame. \begin{wraptable}{r}{0.7\textwidth}
    \vspace{-7mm}
    \centering
    \resizebox{0.98\linewidth}{!}{
    \begin{tabular}{@{}l@{\,\,}|ccc}
    \toprule
    & \!\textbf{W-Temporal Error} $\downarrow$\! & \!\textbf{W-LPIPS} $\downarrow$\! & \!\textbf{W-SSIM} $\uparrow$\!\\ 
    \hline
Spacetime-GS~\cite{spacetimegaussians}                            &                   
1.127 &     0.003803 &    0.989175\\
4DGS~\cite{wu20244d}                                &      \cellcolor{tableyellow}0.780 &    \cellcolor{tableorange}0.002454                &                   
\cellcolor{tableyellow}0.991685 \\
4DRotorGS~\cite{duan20244d}                          &          \cellcolor{tableorange}0.769 &   \cellcolor{tableyellow}0.002653               &                   
\cellcolor{tableorange}0.992745 \\
Grid4D~\cite{grid4d}   &                   
0.840 &                   
0.002689 &                   
0.991504\\
Ours   &                   
\cellcolor{tablered}0.608 &                   
\cellcolor{tablered}0.001806 &                  
\cellcolor{tablered} 0.994581
    \end{tabular}
    }
    \caption{
     \textbf{Temporal error analysis.} We present a quantitative evaluation of our method's performance against existing techniques, utilizing warped error, warped LPIPS, and warped SSIM as key metrics. Results demonstrate the superior performance of our approach. 
    }
    \label{tab:temporal_metrics}
    \vspace{-7mm}
\end{wraptable}
Metrics like W-Temporal Error, W-LPIPS (lower is better), and W-SSIM (higher is better) quantify this consistency. We observe that the renderings from our approach are the most coherent over time compared to the other techniques.

\noindent
\textbf{Evaluation on D-NeRF synthetic} \textbf{dataset.} 
We also evaluated our method against baseline methods on D-NeRF dataset with Single-Scale training\begin{wraptable}{r}{0.6\textwidth}
    \vspace{-7mm}
    \centering
    \resizebox{\linewidth}{!}{
    \begin{tabular}{@{}l@{\,\,}|ccc}
    \toprule
    & \!\textbf{PSNR} $\uparrow$\! & \!\textbf{SSIM} $\uparrow$\! & \!\textbf{LPIPS} $\downarrow$\!\\ 
    \hline
    Deformable3DGS~\cite{yang2024deformable} & \cellcolor{tableorange}38.23 & \cellcolor{tableorange}0.985 & \cellcolor{tableorange}0.017 \\
SARO-GS~\cite{sarogs}          &                   \cellcolor{tableyellow}35.30 &                  0.978 &                   \cellcolor{tableyellow}0.024\\
4DGS~\cite{wu20244d}                                &                   34.19 &                   \cellcolor{tableyellow}0.979 &                   0.026 \\
4DRotorGS~\cite{duan20244d}                          &                   34.22 &                   0.974 &                   \cellcolor{tableyellow}0.024 \\
Grid4D~\cite{grid4d}   &                  \cellcolor{tablered} 39.88 &                  \cellcolor{tablered} 0.987 &                  \cellcolor{tablered} 0.012 \\
Ours   &                   34.82 &                   0.974 &                  0.028


  \!\!\!
    \end{tabular}
    }
    \caption{
    \textbf{Single-scale Training and Same-scale Testing on Synthetic Dataset~\cite{pumarola2021d}.} We observe that the baselines perform better but our method excels in Multi-Scale Evaluation.(Tab.~\ref{tab:avg_synthetic_single_train_multi_test})
    }
    \label{tab:avg_synthetic_results_single_scale}
    \vspace{-10mm}
\end{wraptable} 
and Multi-scale testing setting. Consistent with results on real-world dataset, our method achieved comparable performance at $1\times$ resolution, but demonstrated significant improvements at $2\times$ and $4\times$ resolution. In particular, our method outperformed several baselines, including a $5.06\,dB$ increase over Deformable3DGS, a $5.21\,dB$ increase over SARO-GS, a $3.83\,dB$ increase over Grid4D, a $6.78\,dB$ increase over 4DGS, and a $7.53\,dB$ increase over 4DRotorGS.
Furthermore, the baseline methods exhibited a substantial decline in performance at higher resolutions. For example, Grid4D's PSNR dropped by $10.24\,dB$ from $1\times$  to $4\times$ resolution.

Qualitative results in Fig.~\ref{fig:qual_synthetic} further support our quantitative findings. \begin{wraptable}{r}{0.6\textwidth}
    \vspace{-7mm}
    \centering
    \resizebox{\linewidth}{!}{
    \begin{tabular}{@{}l@{\,\,}|ccc}
    \toprule
    & \!\textbf{PSNR} $\uparrow$\! & \!\textbf{SSIM} $\uparrow$\! & \!\textbf{LPIPS} $\downarrow$\!\\ 
    \hline
    SARO-GS~\cite{sarogs}          &                   
\cellcolor{tablered}32.09 &                   
\cellcolor{tableorange}0.946 &                    
\cellcolor{tablered}0.138\\
Spacetime-GS~\cite{spacetimegaussians}                            &                   
\cellcolor{tableorange}31.64 &                 
\cellcolor{tablered}0.948 &                 
\cellcolor{tableorange}0.139\\
4DGS~\cite{wu20244d}                                &         31.02 &                   
0.935 &                   
0.150 \\
4DRotorGS~\cite{duan20244d}                          &          31.10 &                  
0.937 &                   
0.145 \\
Grid4D~\cite{grid4d}   &                   
30.09 &                   
0.932 &                   
0.148\\
Ours   &                   
\cellcolor{tableyellow}31.43 &                   
\cellcolor{tableyellow}0.943 &                  
\cellcolor{tableyellow}0.140
    \end{tabular}
    }
    \caption{
    \textbf{Single-scale Training and Same-scale Testing on Plenoptic Video Dataset~\cite{li2022neural}.} Our approach is comparable to baseline methods, with only a $0.003$ SSIM difference from the base SARO-GS representation.
    }
    \label{tab:avg_plenoptic_results}
    \vspace{-0.1in}
\end{wraptable}
For example, in the ``hell-warrior'' scene, observe the artifacts present in the SARO-GS output, which are absent in our results. This is significant, as we build on SARO-GS as our base representation. Likewise, the ``lego'' scene exhibits blurry and aliasing artifacts across all baseline methods. Overall, our method effectively preserves high-frequency details, even when trained at a lower resolution.

\noindent
\textbf{Single-Scale Training}
Tab.~\ref{tab:avg_plenoptic_results} presents the results on the Plenoptic Video dataset, where all baselines were trained and evaluated at the same resolution of $1352\times1014$. Our method performed comparably to the other baselines, although it showed a slight performance decrease compared to SARO-GS. This behaviour is expected and consistent with anti-aliasing filters such as Mip-Splatting~\cite{mipsplatting}: by band-limiting each primitive to satisfy the Nyquist criterion, the filter suppresses the high-frequency components that would otherwise overfit the training-resolution views. This is a small trade-off at the training scale for substantially better fidelity at unseen scales. SSIM decreases by only $0.004$ relative to the base SARO-GS representation in Tab.~\ref{tab:avg_synthetic_results_single_scale}.

\begin{figure}[!t]
  \centering
  \includegraphics[width=\linewidth]{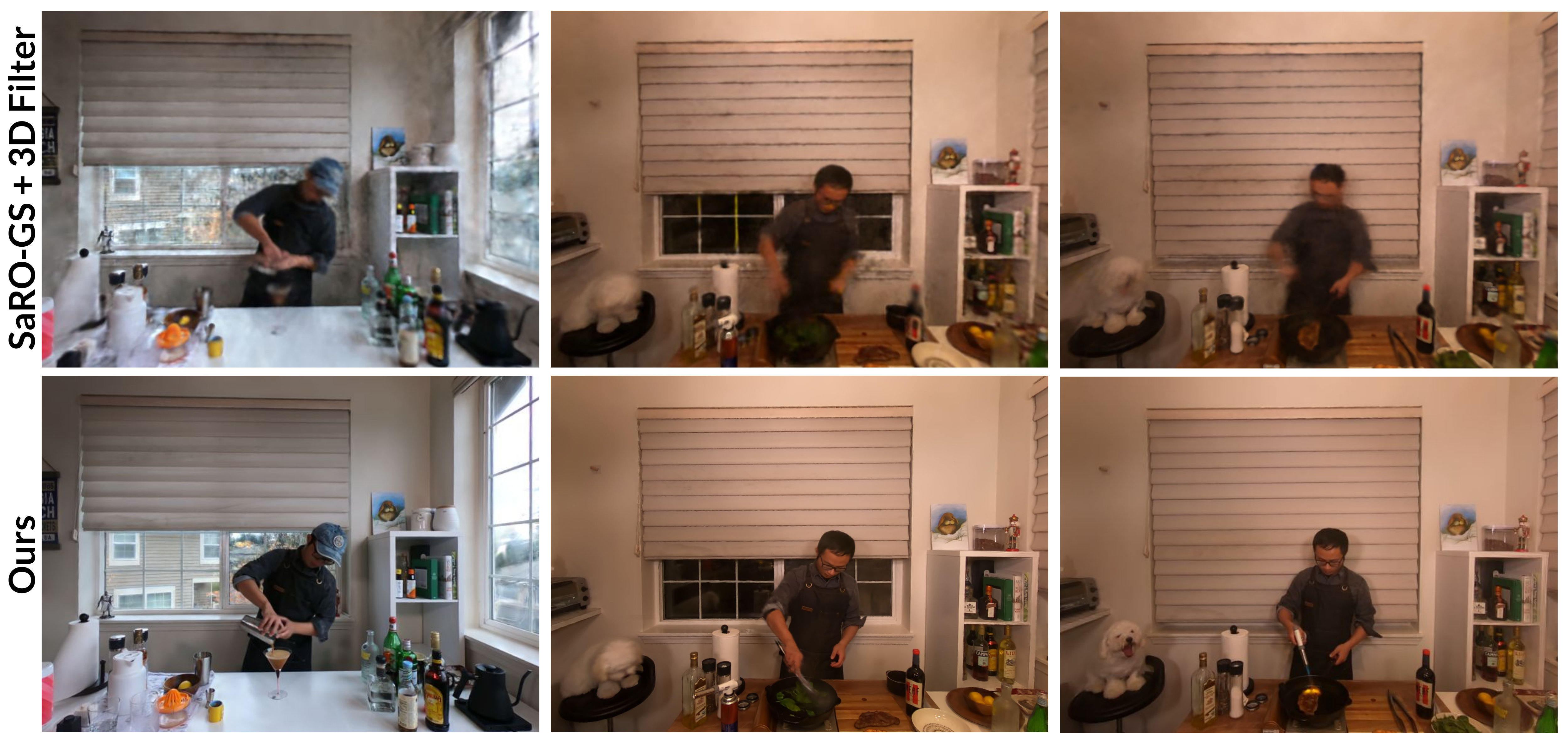}
  \caption{\textbf{Comparison with SARO-GS + static 3D Filter}. Incorporating the static 3D filter from Mip-Splatting~\cite{mipsplatting} into SARO-GS leads to spatial blurring and temporal flickering, especially around the moving human subject. In contrast, our method maintains temporal consistency and preserves high-frequency details.
  }
  \label{fig:qual_saro+3dfilter}
\end{figure}

\noindent
\textbf{Comparison with SARO-GS + static 3D filter.}
To further assess the robustness of our method, we construct an additional baseline by integrating the static 3D filtering module from Mip-Splatting~\cite{mipsplatting} with SARO-GS. \begin{wraptable}{r}{0.6\textwidth}
    \centering
    \resizebox{0.9\linewidth}{!}{
    \begin{tabular}{@{}l@{\,\,}|ccc}
    \toprule
    & \!\textbf{PSNR} $\uparrow$\! & \!\textbf{SSIM} $\uparrow$\! & \!\textbf{LPIPS} $\downarrow$\!\\ 
    \hline
    SARO-GS + 3D Filter          &                        
                              24.92 &   
                              0.840 &   
                              0.270    
                             \\
Ours   &                       
                              \cellcolor{tablered}31.40 &   
                              \cellcolor{tablered}0.941 &   
                              \cellcolor{tablered}0.117   

  \!\!\!
    \end{tabular}
    }
    \vspace{-0.1in}
    \caption{
    \textbf{Quantitative comparison of our method against SARO-GS + 3D Filter}. Our approach significantly outperforms the baseline. This highlights the effectiveness of our method in handling dynamic scenes.}
    \label{tab:ablations}
    \vspace{-0.1in}
\end{wraptable} This baseline aims to simulate the effect of applying spatially consistent 3D filtering to dynamic content. Quantitative results presented in~\cref{tab:ablations} demonstrate that our approach substantially outperforms this baseline across all evaluation metrics (6.4 dB improvement). Further,~\cref{fig:qual_saro+3dfilter} shows that this baseline introduces noticeable blur and temporal inconsistencies, particularly around moving objects such as the human subject. This degradation arises because the static 3D filter fails to account for temporal motion. In contrast, our method effectively preserves fine-grained details and temporal coherence, producing sharper reconstructions and consistent visual quality over time.

\noindent
\textbf{Integration with Other 4D Representations.}
A key advantage of our motion-aware smoothing filter is that it is decoupled from the underlying 4D representation: it operates on the projected 3D Gaussians and merely replaces the constant sampling rate $\hat{\nu}_k$ with a time-varying $\hat{\nu}_k(t)$. To demonstrate this, we integrate it with SpeeDe3DGS~\cite{speede3dgs}, a recent and architecturally different 4D representation. Tab.~\ref{tab:speede3dgs} reports single-scale training and $4\times$ testing on the D-NeRF dataset~\cite{pumarola2021d}. Our filter improves SpeeDe3DGS by $4.45$\,dB in PSNR while using fewer Gaussian primitives, confirming that the gains are not specific to SARO-GS. The overhead is modest: training increases by about two minutes, and the per-primitive density estimate ($\mathcal{O}(NDT)$) adds only $17.7$\,MB of memory; compressing this is left for future work. The lower part of Tab.~\ref{tab:speede3dgs} ablates the kernel bandwidths $h_x, h_t$ and the temporal grid size $T$, showing that our method is robust to these choices. Unless stated otherwise, we use $h_x=0.3$, $h_t=0.05$, $T=100$, and Weibull shape $k=10$ in all experiments.

\begin{table}[!t]
    \centering
    \caption{\textbf{Integration with SpeeDe3DGS~\cite{speede3dgs}} on the D-NeRF dataset~\cite{pumarola2021d} (single-scale training, $4\times$ testing). Our motion-aware filter improves SpeeDe3DGS by $4.45$\,dB while using fewer primitives and adding only $\sim$2 minutes of training. Lower rows ablate the bandwidths $h_x, h_t$ and grid size $T$ ($\dagger$ marks the default configuration). $\#$G is the number of Gaussian primitives.}
    \label{tab:speede3dgs}
    \resizebox{0.82\linewidth}{!}{
    \begin{tabular}{ll|ccccc}
    \toprule
    & & \textbf{PSNR}$\uparrow$ & \textbf{SSIM}$\uparrow$ & \textbf{LPIPS}$\downarrow$ & \textbf{Train (min)} & $\#$\textbf{G} \\
    \midrule
    \multicolumn{2}{l|}{SpeeDe3DGS~\cite{speede3dgs}} & 27.83 & 0.940 & 0.067 & \textbf{4.45} & 2508 \\
    \multicolumn{2}{l|}{\textbf{+ Ours} ($\dagger$)} & \textbf{32.28} & \textbf{0.956} & \textbf{0.064} & 6.48 & \textbf{2330} \\
    \midrule
    \multirow{3}{*}{$h_x$} & 0.1 & 32.13 & 0.955 & 0.065 & 6.41 & 2314 \\
     & 0.2 & 32.20 & 0.956 & 0.064 & 6.31 & 2312 \\
     & 0.3$^\dagger$ & 32.28 & 0.956 & 0.064 & 6.48 & 2330 \\
    \midrule
    \multirow{3}{*}{$h_t$} & 0.05$^\dagger$ & 32.28 & 0.956 & 0.064 & 6.48 & 2330 \\
     & 0.1 & 32.13 & 0.956 & 0.065 & 6.47 & 2339 \\
     & 0.2 & 32.07 & 0.956 & 0.065 & 6.47 & 2352 \\
    \midrule
    \multirow{2}{*}{$T$} & 50 & 32.25 & 0.956 & 0.064 & 6.41 & 2290 \\
     & 100$^\dagger$ & 32.28 & 0.956 & 0.064 & 6.48 & 2330 \\
    \bottomrule
    \end{tabular}
    }
\end{table}

\noindent
\textbf{Zoom-Out (Sub-$1\times$) Evaluation.}
The evaluations before are for the resolutions above the training resolution (zoom-in). We now assess the opposite regime, rendering \emph{below} it. Using SpeeDe3DGS on the D-NeRF dataset, we train at $1\times$ ($800\times800$) and render at $1\times$, $\tfrac{1}{2}\times$ ($400\times400$), and $\tfrac{1}{4}\times$ ($200\times200$). As shown in Tab.~\ref{tab:zoomout}, the two methods are on par at the training resolution, but as the render resolution decreases the base representation degrades sharply---its PSNR falls from $35.0$ to $28.1$\,dB at $\tfrac{1}{4}\times$---whereas our motion-aware filter stays nearly constant (${\sim}35.9$\,dB) across all scales, i.e., it is scale-consistent in both directions. We note that the large PSNR gap at $\tfrac{1}{4}\times$ is concentrated at high-frequency regions (check analysis in supplementary material).

\begin{table}[!t]
    \centering
    \caption{\textbf{Zoom-out (sub-$1\times$) evaluation with SpeeDe3DGS~\cite{speede3dgs} on the D-NeRF dataset~\cite{pumarola2021d}} (averaged over 8 scenes). All models are trained at $1\times$ ($800\times800$) and rendered at lower resolutions. Our filter keeps quality nearly flat across scales, whereas the base degrades as the render resolution drops.}
    \label{tab:zoomout}
    \resizebox{0.85\linewidth}{!}{
    \begin{tabular}{l|cc|cc|cc}
    \toprule
    & \multicolumn{2}{c|}{PSNR $\uparrow$} & \multicolumn{2}{c|}{SSIM $\uparrow$} & \multicolumn{2}{c}{LPIPS $\downarrow$} \\
    Eval resolution & SpeeDe3DGS & \textbf{+Ours} & SpeeDe3DGS & \textbf{+Ours} & SpeeDe3DGS & \textbf{+Ours} \\
    \midrule
    $1\times$ ($800\times800$) & 35.01 & 35.02 & 0.974 & 0.974 & 0.040 & 0.041 \\
    $\tfrac{1}{2}\times$ ($400\times400$) & 32.89 & \textbf{35.88} & 0.975 & \textbf{0.981} & 0.025 & \textbf{0.024} \\
    $\tfrac{1}{4}\times$ ($200\times200$) & 28.10 & \textbf{35.87} & 0.951 & \textbf{0.985} & 0.029 & \textbf{0.014} \\
    \bottomrule
    \end{tabular}
    }
\end{table}

\section{Limitations}
The main limitation of our method is its memory footprint. For each Gaussian primitive, we store a discrete density matrix over the joint $(f/d,\,t)$ grid; with $N$ primitives and a grid of size $D\times T$, where $D$ and $T$ are the numbers of discretization bins along $f/d$ and time respectively, the total memory scales as $\mathcal{O}(N \times D \times T)$. This overhead grows with the grid resolution and the number of primitives, which can become significant for large scenes. Several directions can mitigate this in future work. Since inference only requires the per-timestamp maximum of the conditional density, the full joint matrix can be discarded and replaced by the resulting time-varying sampling-rate curve, reducing the footprint from $\mathcal{O}(NDT)$ to $\mathcal{O}(NT)$. Another approach can be that a small shared network predicts the sampling rate directly from a primitive's features and the query time, removing the need to store a matrix per primitive. Finally, primitives undergoing similar motion can be clustered to share a single density estimate. We leave these memory-efficient extensions to future work.

\section{Conclusion and Future Work}
We present a motion-aware smoothing filter designed to achieve alias-free rendering for 4D representations, enabling effective modelling of dynamic scenes. Our filter estimates the appropriate sampling interval by estimating a joint distribution and, during inference, selects the value corresponding to the maximum probability for a given query timestamp. Experimental results demonstrate that the proposed method outperforms baseline 4D representations when trained at a lower resolution and evaluated at higher resolutions. This quantitative and qualitative analysis further corroborates our claim that this smoothing filter excels at zoom-in/zoom-out rendering. Furthermore, this strategy is flexible and can be applied to any 4D representation. In future work, we will explore its application to more challenging dynamic scenes characterized by rapid, complex motion, pushing the boundaries of high-fidelity $4D$ reconstruction for per-scene methods and feed-forward methods.

\section*{Acknowledgement}
This project
is supported by Google and Kotak IISc AI-ML Centre
(KIAC). Ankit Dhiman is supported by Samsung R\&D Institute India - Bangalore. 

\bibliographystyle{splncs04}
\bibliography{main}

\clearpage

\newcommand{\supptitle}{Towards Alias-Free 4D Gaussian Representations with Motion-Aware Filtering}

\newcommand{\maketitlesupplementary}{%
  \begin{center}
    {\Large\bfseries\boldmath
     \pretolerance=10000
     \supptitle\par}%
    \vskip .8cm
    {\large Supplementary Material\par}%
  \end{center}%
  \vskip .5cm
}

\maketitlesupplementary

\appendix

\makeatletter
\providecommand*{\authcount}[1]{}
\renewcommand*{\l@title}[2]{}
\renewcommand*{\l@author}[2]{}
\section*{\contentsname}
\@starttoc{toc}
\makeatother
\addtocontents{toc}{\protect\setcounter{tocdepth}{2}}

\section{Additional Details}
\label{sec:supp_additional_details}

\subsection{How f/d graph was obtained?}
\label{sec:supp_df_dist_graph_exp}

\begin{wrapfigure}{r}{0.45\textwidth}
    \centering
    \vspace{-25pt}
    \includegraphics[width=\linewidth]{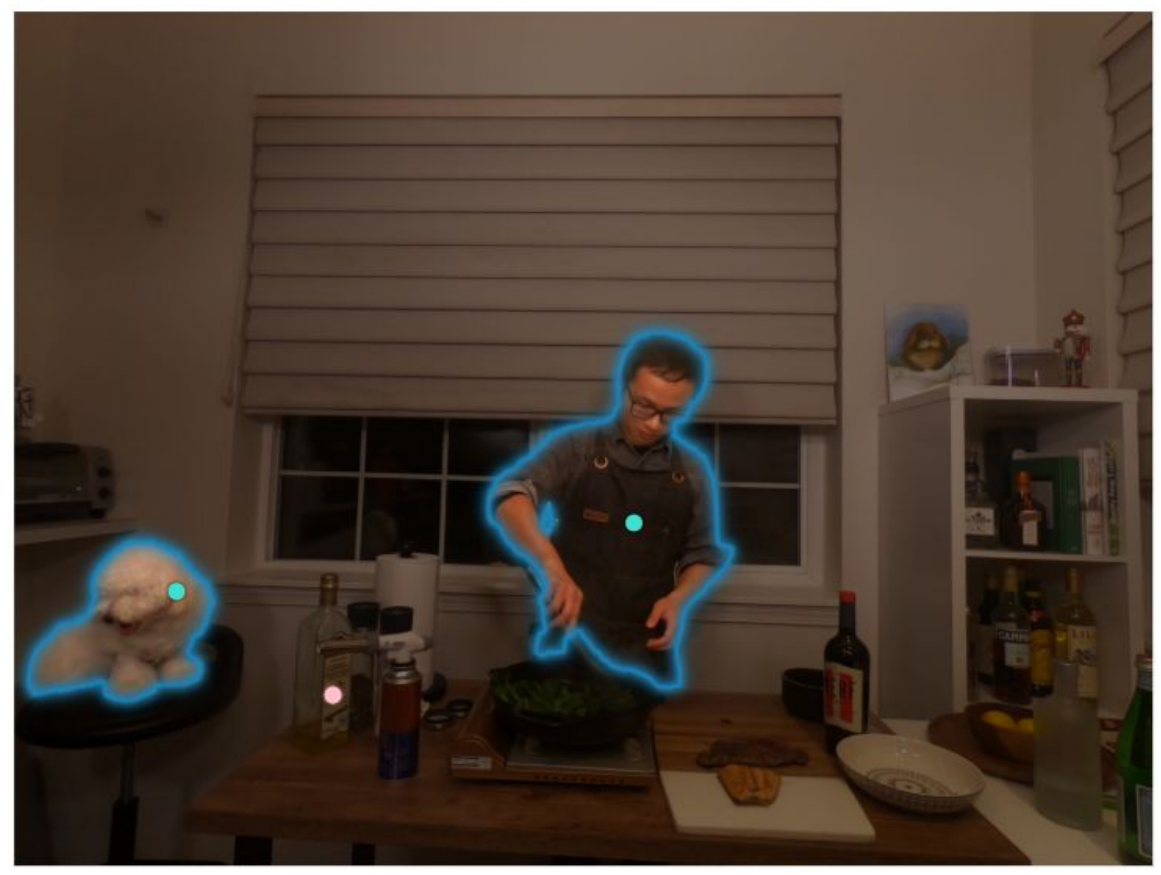}
    \caption{To segment the dynamic region in the image, we utilize SAM~\cite{sam}, an online interactive tool. By providing positive clicks, we identify the relevant masks, and any outliers are removed with negative clicks.}
    \label{fig:supp_d_f_analysis_fig}
    \vspace{-15pt}
\end{wrapfigure} 
To identify the dynamic region in a scene, we project Gaussian primitives onto an image and utilize SAM~\cite{sam} with manually provided seed points to obtain static and dynamic masks (see Fig.~\ref{fig:supp_d_f_analysis_fig} ). We then classify the Gaussians that fall within the static region as static, and those within the dynamic region as dynamic. This classification allows us to separate static and dynamic primitives from each other and we provide the analysis for the classified primitives in \cref{fig:dist_d/f}.

\subsection{Temporal Stability Metrics}
We use the metrics described in~\cite{lai2018learning} to measure the output videos' temporal stability and perceptual similarity. Metrics used are as follows:

\paragraph{Warped Temporal Error (W-Temporal Error)}
To measure a video's temporal stability, we use the flow warping error, which quantifies the flow warping error between two frames:
\begin{equation}
	\mathcal{E}_{\text{warp}}(I_t, I_{t+1}) = \frac{1}{N} \sum_{i=1}^N  \| I_t^{(i)} - \hat{I}_{t+1}^{(i)} \|_2^2, 
\end{equation}
We measure the warping error between two frames by calculating the difference between a warped frame $\hat{I}_{t+1}$ from $I_{t+1}$ and previous frame $I_t$. To warp $I_t$ using optical flow to $I_{t+1}$, we initialize the warped frame as $I_t$, then move pixels according to the flow, overwriting target locations and leaving occluded regions with their original values from $I_t$. The warped temporal error of a video, $V$, is calculated as:
\begin{equation}
	W-Temporal\,Error(V) =  \sum_{t=1}^{T-1} \mathcal{E}_{\text{warp}}(I_t, I_{t+1}), 
\end{equation}

\paragraph{Warped LPIPS}
Further, on the similar lines of the warped error between two frames, we define warped LPIPS~\cite{zhang2018unreasonable} between two frames as follows:
\begin{equation}
	{LPIPS}_{\text{warp}}(I_t, I_{t+1}) = LPIPS( I_t^{(i)},\hat{I}_{t+1}^{(i)})
\end{equation}
where $\hat{I}_{t+1}$ is the warped frame as defined in the previous paragraph. The warped LPIPS of a video, $V$, is calculated as:
\begin{equation}
	W-LPIPS(V) =  \frac{1}{T-1} \sum_{t=1}^{T-1} {LPIPS}_{\text{warp}}(I_t, I_{t+1}), 
\end{equation}

\paragraph{Warped SSIM}
Similar to the warped LPIPS, we compute warped SSIM. 
\begin{equation}
	{SSIM}_{\text{warp}}(I_t, I_{t+1}) = SSIM( I_t^{(i)},\hat{I}_{t+1}^{(i)})
\end{equation}
where $\hat{I}_{t+1}$ is the warped frame as defined in the previous paragraph. The warped LPIPS of a video, $V$, is calculated as:
\begin{equation}
	W-SSIM(V) =  \frac{1}{T-1} \sum_{t=1}^{T-1} {SSIM}_{\text{warp}}(I_t, I_{t+1}), 
\end{equation}

\noindent We use RAFT \cite{teed2020raftrecurrentallpairsfield} to calculate optical flow in all temporal stability metrics.

\subsection{Information on Weibull Distribution}
The probability density function of the Weibull distribution  is given by:
\[
f(x; k, \lambda) =
\frac{k}{\lambda} \left(\frac{x}{\lambda}\right)^{k-1} e^{-(x/\lambda)^k}, \quad x > 0
\]

where:
\begin{itemize}
    \item \( k \) is the \textbf{shape parameter},
    \item \( \lambda \) is the \textbf{scale parameter}.
\end{itemize}

\subsection{Ablation Studies}
\label{sec:supp_ablations}
We ablate the two key hyperparameters that govern the temporal smoothing of our motion-aware filter, using the SARO-GS representation on the D-NeRF dataset~\cite{pumarola2021d}: the temporal grid size $T$ (Tab.~\ref{tab:supp_abl_grid}) and the interval, in training iterations, at which the filter is recomputed (Tab.~\ref{tab:supp_abl_update}). For the grid size, performance peaks at $T=100$ across all scales: too few bins ($T=25$) under-resolve the temporal variation of $f/d$, whereas too many ($T=150$) make the per-bin density estimates noisy. For the update interval, recomputing the filter more frequently (every $100$ iterations) performs best, as the density estimate then tracks the evolving primitive geometry more closely. These trends justify our defaults of $T=100$ and a $100$-iteration update interval. The kernel bandwidths $h_x$ and $h_t$ are additionally ablated on SpeeDe3DGS~\cite{speede3dgs} in \cref{tab:speede3dgs}.

\begin{table}[!t]
    \centering
    \caption{\textbf{Ablation on the temporal grid size $T$} (SARO-GS, D-NeRF~\cite{pumarola2021d}, single-scale training and multi-scale testing). Our default $T=100$ (bold) performs best across all scales.}
    \label{tab:supp_abl_grid}
    \begin{tabular}{c|c|cccc}
    \toprule
    \multicolumn{2}{c|}{Grid Size ($T$)} & 25 & 50 & \textbf{100} & 150 \\
    \midrule
    \multirow{3}{*}{PSNR $\uparrow$}
     & $1\times$ & 27.88 & 28.44 & \textbf{28.97} & 28.32 \\
     & $2\times$ & 28.93 & 29.54 & \textbf{30.05} & 29.41 \\
     & $4\times$ & 30.58 & 31.28 & \textbf{31.72} & 31.11 \\
    \midrule
    \multirow{3}{*}{LPIPS $\downarrow$}
     & $1\times$ & 0.060 & 0.058 & \textbf{0.056} & 0.059 \\
     & $2\times$ & 0.049 & 0.045 & \textbf{0.041} & 0.046 \\
     & $4\times$ & 0.043 & 0.037 & \textbf{0.029} & 0.038 \\
    \bottomrule
    \end{tabular}
\end{table}

\begin{table}[!t]
    \centering
    \caption{\textbf{Ablation on the filter update interval} (in training iterations; SARO-GS, D-NeRF~\cite{pumarola2021d}). Updating every $100$ iterations (bold, our default) performs best.}
    \label{tab:supp_abl_update}
    \begin{tabular}{c|c|ccc}
    \toprule
    \multicolumn{2}{c|}{Update Interval} & \textbf{100} & 200 & 300 \\
    \midrule
    \multirow{3}{*}{PSNR $\uparrow$}
     & $1\times$ & \textbf{28.97} & 27.97 & 26.83 \\
     & $2\times$ & \textbf{30.05} & 29.01 & 27.81 \\
     & $4\times$ & \textbf{31.72} & 30.64 & 29.34 \\
    \bottomrule
    \end{tabular}
\end{table}

\section{Additional Results}
\label{sec:supp_additional_results}

\subsection{Per-Scene Zoom-Out Results}
\label{sec:supp_zoomout}
Tab.~\ref{tab:supp_zoomout} reports the per-scene PSNR for the zoom-out (sub-$1\times$) evaluation summarized in \cref{tab:zoomout}, for all eight D-NeRF~\cite{pumarola2021d} scenes. Models are trained at $1\times$ ($800\times800$) and rendered at $1\times$, $\tfrac{1}{2}\times$ ($400\times400$), and $\tfrac{1}{4}\times$ ($200\times200$). Every scene follows the same trend: the base (SpeeDe3DGS) and our filter are on par at the training resolution, but the base degrades as the render resolution drops while our method stays stable, with the largest gains on scenes containing thin structures (\eg, mutant, trex).

\begin{table}[!t]
    \centering
    \caption{\textbf{Per-scene PSNR for the zoom-out (sub-$1\times$) evaluation} on the D-NeRF dataset~\cite{pumarola2021d}. B = SpeeDe3DGS, O = +Ours.}
    \label{tab:supp_zoomout}
    \resizebox{0.9\linewidth}{!}{
    \begin{tabular}{l|cc|cc|cc}
    \toprule
    & \multicolumn{2}{c|}{$1\times$ ($800\times800$)} & \multicolumn{2}{c|}{$\tfrac{1}{2}\times$ ($400\times400$)} & \multicolumn{2}{c}{$\tfrac{1}{4}\times$ ($200\times200$)} \\
    Scene & B & O & B & O & B & O \\
    \midrule
    bouncing balls & 38.08 & 38.71 & 30.64 & 37.14 & 23.89 & 33.06 \\
    hell warrior & 38.84 & 38.87 & 39.19 & 39.72 & 37.19 & 40.94 \\
    hook & 33.61 & 33.55 & 33.13 & 34.71 & 28.97 & 35.84 \\
    jumping jacks & 34.97 & 34.84 & 34.71 & 36.11 & 29.74 & 37.23 \\
    lego & 24.65 & 24.66 & 24.33 & 25.20 & 22.37 & 25.82 \\
    mutant & 36.85 & 36.75 & 33.93 & 38.87 & 27.31 & 39.20 \\
    stand up & 38.60 & 38.37 & 36.98 & 39.69 & 31.19 & 40.10 \\
    trex & 34.50 & 34.44 & 30.19 & 35.64 & 24.12 & 34.77 \\
    \midrule
    Average & 35.01 & 35.02 & 32.89 & \textbf{35.88} & 28.10 & \textbf{35.87} \\
    \bottomrule
    \end{tabular}
    }
\end{table}

\begin{figure}[!t]
    \centering
    \includegraphics[width=\linewidth]{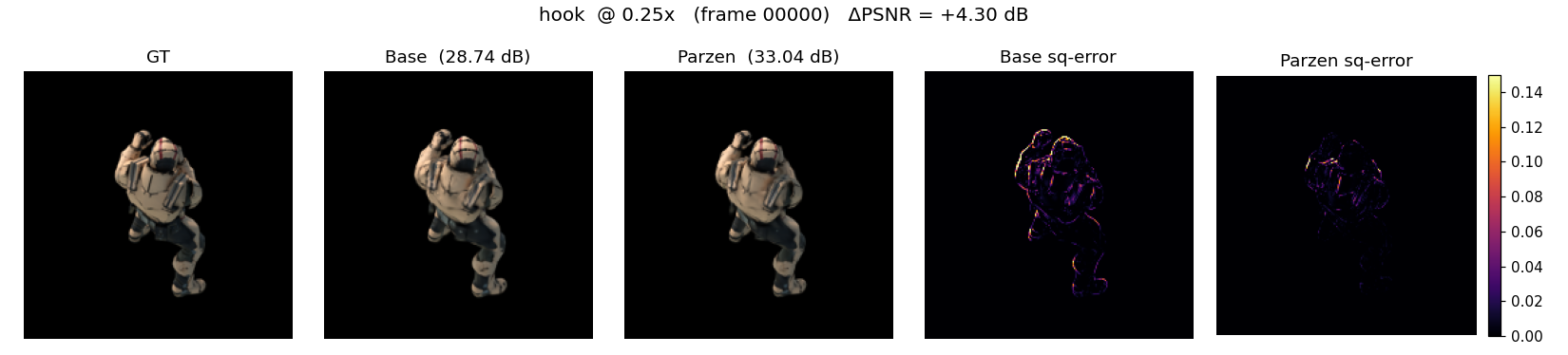}\\[2pt]
    \includegraphics[width=\linewidth]{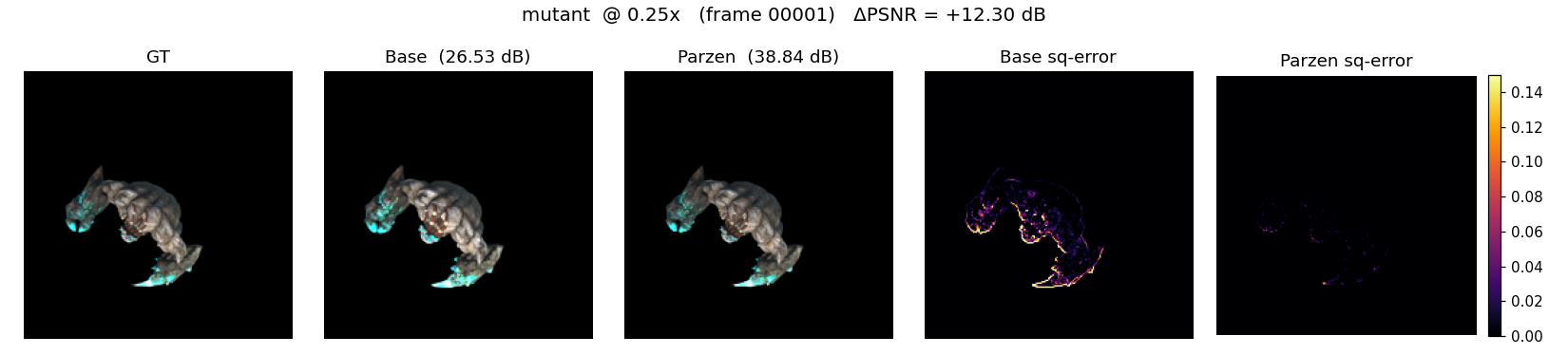}
    \caption{\textbf{Zoom-out ($\tfrac{1}{4}\times$) error maps} for the \emph{hook} (top) and \emph{mutant} (bottom) scenes. ``Base'' is SpeeDe3DGS and ``Parzen'' is our filter. Left to right: ground truth, base render, our render, and the two squared-error maps (shared scale). The base and our renders are visually near-identical, yet the base's error is concentrated along object silhouettes and edges (bright), which our filter removes (near-dark error map). This explains why the large PSNR gains (e.g., $+12.3$\,dB on \emph{mutant}) are perceptually subtle: they arise almost entirely from a thin ring of aliased edge pixels.}
    \label{fig:supp_zoomout_errmaps}
\end{figure}

\noindent
\textbf{Where the Zoom-Out Gains Come From.}
The sub-$1\times$ PSNR improvements in \cref{tab:zoomout} are substantial (up to $+7.8$\,dB at $\tfrac{1}{4}\times$), yet the rendered images look nearly identical by eye. We reconcile this by analyzing \emph{where} the error reduction occurs. Two observations show that the gain is concentrated in a small set of pixels: (i)~the base and our renders differ by only ${\sim}0.9\%$ in mean absolute pixel value, and (ii)~at $\tfrac{1}{4}\times$, $3.5\%$ of pixels account for $90\%$ of the base representation's error, and the fraction of high-error pixels (squared error $>0.05$) drops from $1.35\%$ (base) to $0.03\%$ (ours). This is visualized in Fig.~\ref{fig:supp_zoomout_errmaps}: the base render's error is concentrated along object silhouettes and edges, which our filter removes.

To localize the gain, we partition each test image into rings by distance from the object silhouette and measure the share of the total Base$\rightarrow$Ours error reduction contributed by each ring, averaged over all $8$ scenes and frames (Tab.~\ref{tab:supp_region}). At both $\tfrac{1}{4}\times$ and $\tfrac{1}{2}\times$, the silhouette boundary band ($\le 2$\,px on either side of the object edge) accounts for ${\sim}70\%$ of the entire error reduction while covering only ${\sim}7\%$ of the pixels; the far background contributes essentially nothing. Even within the genuine interior ($>5$\,px deep), the reduction is dominated by high-frequency texture edges ($56.8\%$ of the interior gain at $\tfrac{1}{4}\times$), with only a modest contribution from flat regions.

In short, our filter performs \emph{anti-aliasing of high-frequency edges}---silhouettes first, interior texture second---rather than a broad reconstruction-quality improvement. This is why the large PSNR gains, driven by the sensitivity of MSE to a thin ring of aliased edge pixels, are perceptually subtle, and why the structural and perceptual metrics move far less (at $\tfrac{1}{4}\times$, SSIM $0.951\!\rightarrow\!0.985$ and LPIPS $0.029\!\rightarrow\!0.014$). We therefore report PSNR together with SSIM and LPIPS for the zoom-out setting.

\begin{table}[!t]
    \centering
    \caption{\textbf{Where the zoom-out error reduction comes from.} Share of the total Base$\rightarrow$Ours PSNR error reduction by distance from the object silhouette (D-NeRF, 8 scenes, all frames). The boundary band dominates while covering few pixels; the far background contributes almost nothing.}
    \label{tab:supp_region}
    \begin{tabular}{l|cc}
    \toprule
    Region (distance from silhouette) & \multicolumn{2}{c}{\% of error reduction} \\
     & $\tfrac{1}{4}\times$ & $\tfrac{1}{2}\times$ \\
    \midrule
    Silhouette ring ($d{=}1$\,px, ${\sim}1.7\%$ of pixels) & 50.7 & 57.8 \\
    Inside band ($d{=}2$--$5$\,px) & 11.8 & 6.2 \\
    Deep interior ($>5$\,px) & 20.3 & 25.8 \\
    Background halo ($\le 2$\,px) & 17.0 & 9.4 \\
    Far background (${\sim}77\%$ of pixels) & 0.2 & 0.7 \\
    \bottomrule
    \end{tabular}
\end{table}

\subsection{Per-Scene Results}
\label{sec:supp_per_scene_all}
\begin{figure}[!t]
    \centering
    \includegraphics[width=0.9\linewidth]{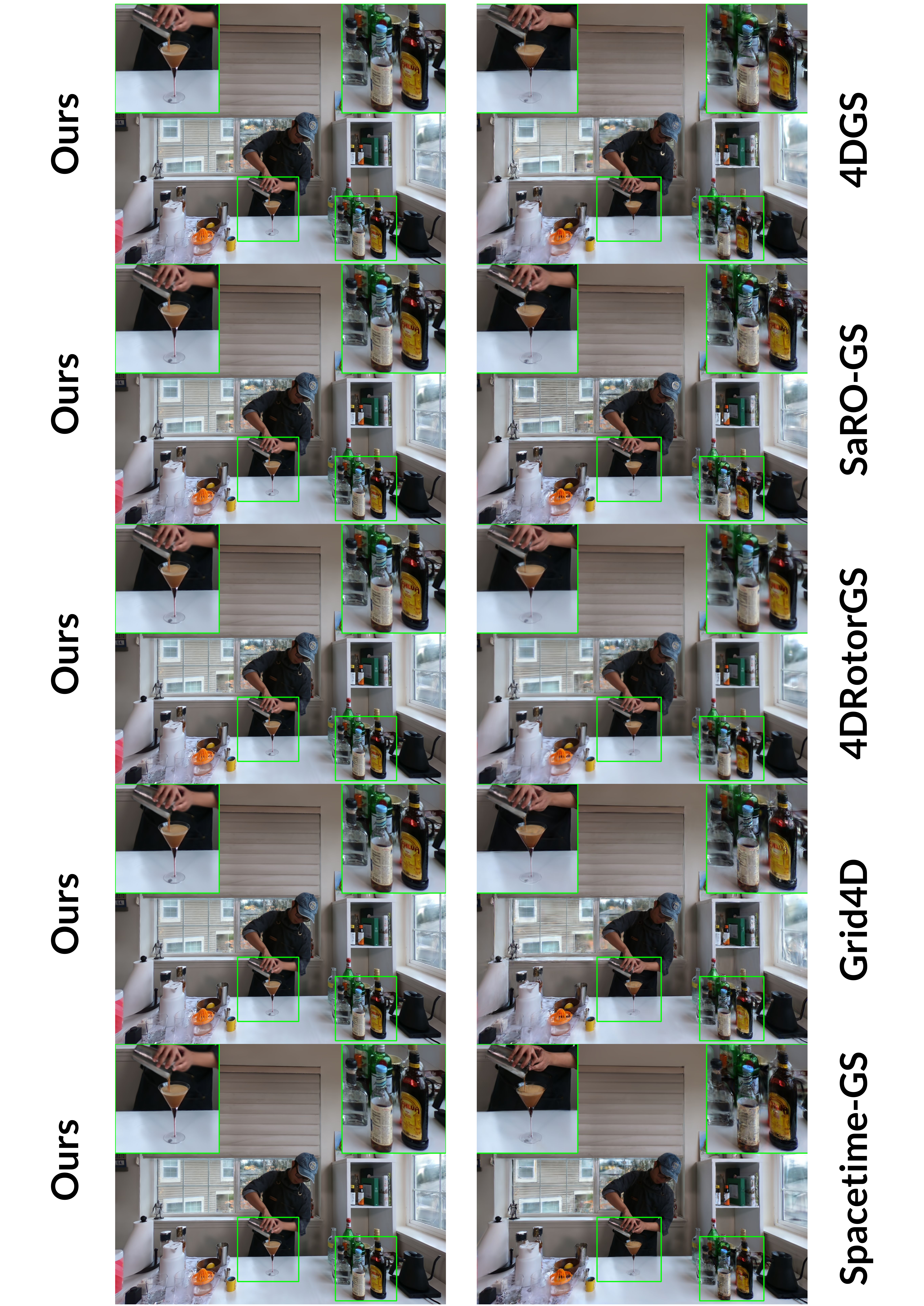}
    \caption{More Qualitative comparisons for coffee martini scene of plenoptic dataset. The differences can be observed in the flow of the coffee from the steel glass to the martini glass for dynamic region \& for the static regions the things kept on the left \& right side of the table.}
    \label{fig:supp_qual_plenoptic_coffeemartini}
\end{figure}

\begin{figure}[!t]
    \centering
    \includegraphics[width=0.9\linewidth]{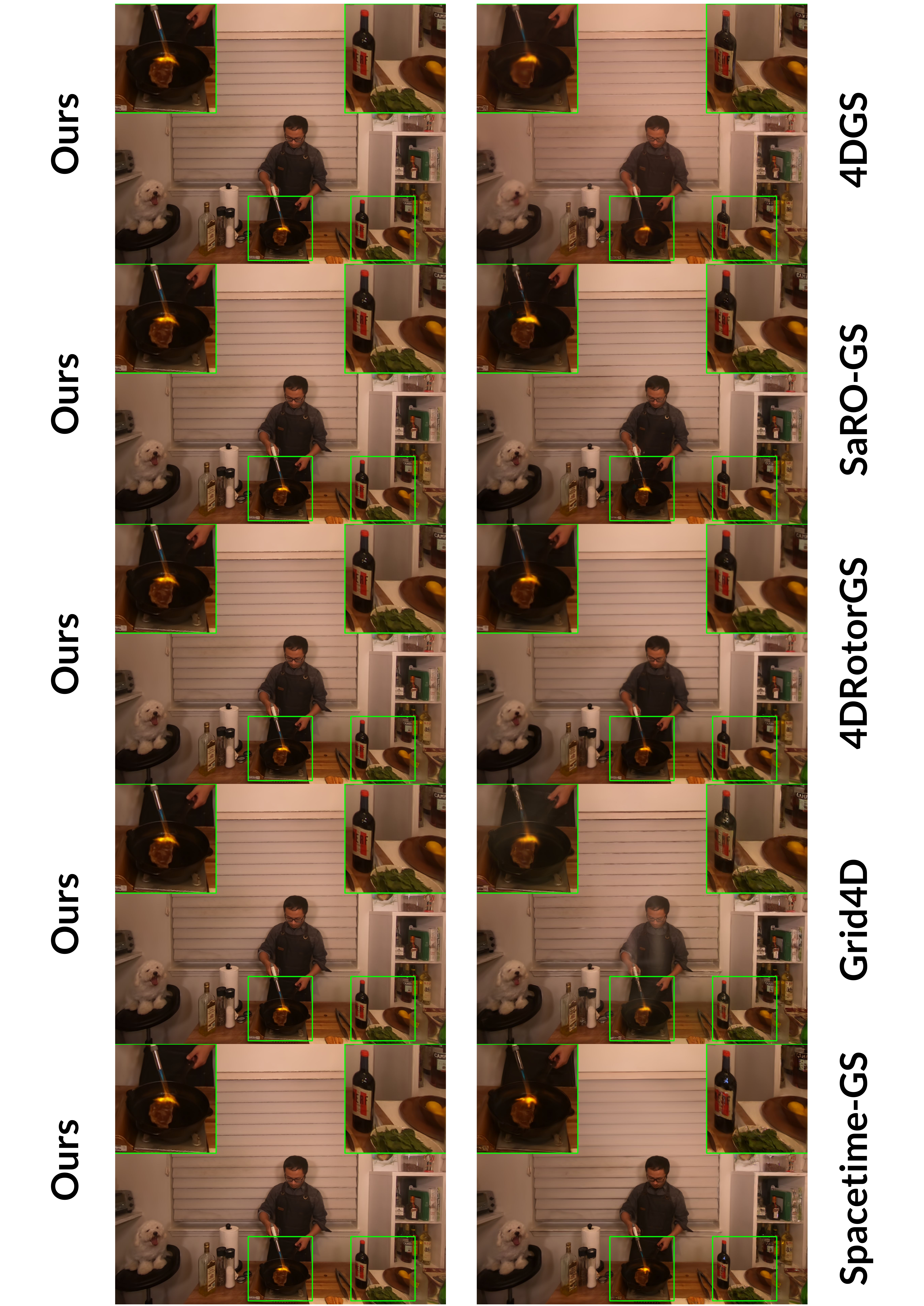}
    \caption{More Qualitative comparisons for flame steak scene of plenoptic dataset. The differences can be observed at tip of the torch gun \& the flame coming out of it for dynamic region \& for static region the blinds \& the things kept on the left \& right side of the table.}
    \label{fig:supp_qual_plenoptic_1}
\end{figure}

\begin{figure}[!t]
    \centering
    \includegraphics[width=0.9\linewidth]{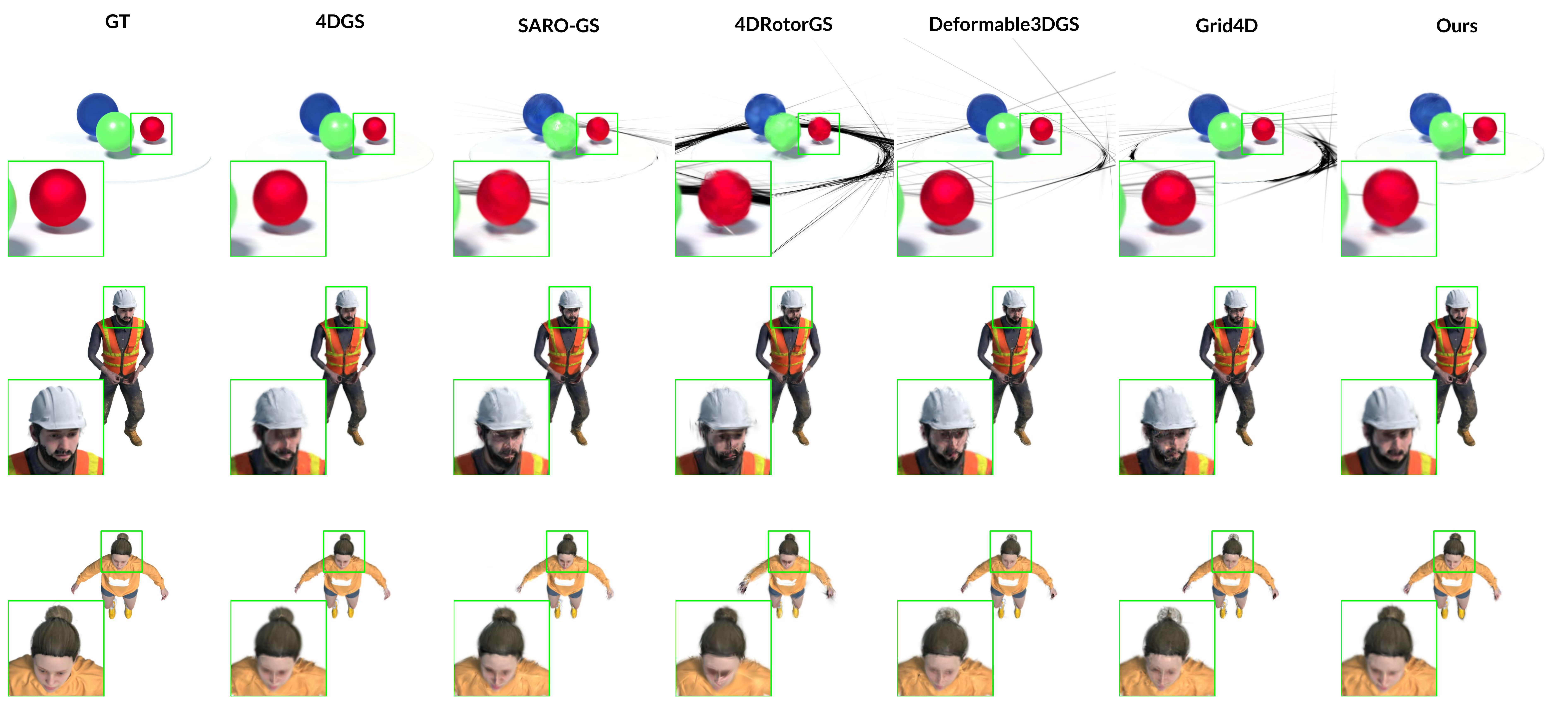}
    \caption{More qualitative results on synthetic dataset}
    \label{fig:supp_qual_plenoptic_2}
\end{figure}

We show per-scene quantitative results for Plenoptic Video~\cite{li2022neural} dataset in Tabs.~\ref{tab:avg_plenoptic_single_train_multi_test_coffeemartini},~\ref{tab:avg_plenoptic_single_train_multi_test_cookspinach},\\~\ref{tab:avg_plenoptic_single_train_multi_test_cutroastedbeef},~\ref{tab:avg_plenoptic_single_train_multi_test_flamesalmon},~\ref{tab:avg_plenoptic_single_train_multi_test_flamesteak},~\ref{tab:avg_plenoptic_single_train_multi_test_searsteak}. Further, we show additional qualitative results in Figs. ~\ref{fig:supp_qual_plenoptic_coffeemartini},~\ref{fig:supp_qual_plenoptic_1},~\ref{fig:supp_qual_plenoptic_2}.  
We show scenewise temporal metrics results in Tabs.~\ref{tab:temporal_plenoptic_searsteak}, ~\ref{tab:temporal_plenoptic_flameteak} and ~\ref{tab:temporal_plenoptic_flamesalmon}

\begin{table}[]
    \renewcommand{\tabcolsep}{1pt}
    
    \centering
    \caption{
    \textbf{Single-Scale Training and Multi-Scale Evaluation on cook spinach scene of Plenoptic Video Dataset~\cite{li2022neural}.} All methods are trained at the smallest scale ($1 \times$) that is $676$x$507$ and evaluated across multiple scales ($1 \times$, $2 \times$, and $4 \times$), where increasing resolution simulates zoom-in effects. While our approach performs comparably to existing methods at the training resolution($1 \times$), it demonstrates superior fidelity across all other scales, significantly outperforming prior work.}  
    \resizebox{0.95\linewidth}{!}{
    \begin{tabular}{@{}l@{\,\,}|cccc|cccc|cccc}
    \toprule
    & \multicolumn{4}{c|}{PSNR $\uparrow$} & \multicolumn{4}{c|}{SSIM $\uparrow$} & \multicolumn{4}{c}{LPIPS $\downarrow$}  \\
    & $1 \times$ Res. & $2 \times$ Res. & $4 \times$ Res.  & Avg. & $1 \times$ Res. & $2 \times$ Res. & $4 \times$ Res. &  Avg. & $1 \times$ Res. & $2 \times$ Res. & $4 \times$ Res. &  Avg.  \\ \hline
    SARO-GS~\cite{sarogs}          &                        
                              34.01 &   
                              30.46 &
                              28.34 &
                              30.93 &
                              0.962 &   
                              0.927 &
                              0.902 &
                              0.930 &
                              0.084 &   
                              0.156 &
                              0.222 &
                              0.154 \\
Spacetime-GS~\cite{spacetimegaussians}                            & 
                              34.24 &   
                              30.0 &
                              27.56 &
                              30.6 &
                              0.964 &   
                              0.928 &
                              0.903 &
                              0.931 &
                              0.085 &   
                              0.165 &
                              0.224 &
                              0.158 \\
4DGS~\cite{wu20244d}                                &                       
                              33.61 &   
                              29.75 &
                              27.46 &
                              30.27 &
                              0.955 &   
                              0.916 &
                              0.896 &
                              0.922 &
                              0.095 &   
                              0.177 &
                              0.230 &
                              0.167 \\
4DRotorGS~\cite{duan20244d}                          &                       
                              34.26 &   
                              31.19 &
                              30.79 &
                              32.08 &
                              0.966 &   
                              0.933 &
                              0.940 &
                              0.946 &
                              0.085 &   
                              0.187 &
                              0.226 &
                              0.166 \\
Grid4D~\cite{grid4d}   &                       
                              33.02 &   
                              29.34 &
                              27.66 &
                              30.0 &
                              0.955 &   
                              0.912 &
                              0.892 &
                              0.919 &
                              0.091 &   
                              0.174 &
                              0.228 &
                              0.164  \\
Ours   &                       
                              33.30 &   
                              31.85 &
                              31.02 &
                              32.05 &
                              0.957 &   
                              0.939 &
                              0.925 &
                              0.940 &
                              0.092 &   
                              0.159 &
                              0.222 &
                              0.157 
    \end{tabular}
    }
     \label{tab:avg_plenoptic_single_train_multi_test_cookspinach}
\end{table}

\begin{table}[!t]
    \renewcommand{\tabcolsep}{1pt}
    \centering
    \caption{
    \textbf{Single-Scale Training and Multi-Scale Evaluation on coffee martini scene of Plenoptic Video Dataset~\cite{li2022neural}.} All methods are trained at the smallest scale ($1 \times$) that is $676$x$507$ and evaluated across multiple scales ($1 \times$, $2 \times$, and $4 \times$), where increasing resolution simulates zoom-in effects. While our approach performs comparably to existing methods at the training resolution($1 \times$), it demonstrates superior fidelity across all other scales, significantly outperforming prior work.} 
    \resizebox{0.95\linewidth}{!}{
    \begin{tabular}{@{}l@{\,\,}|cccc|cccc|cccc}
    \toprule
    & \multicolumn{4}{c|}{PSNR $\uparrow$} & \multicolumn{4}{c|}{SSIM $\uparrow$} & \multicolumn{4}{c}{LPIPS $\downarrow$}  \\
    & $1 \times$ Res. & $2 \times$ Res. & $4 \times$ Res.  & Avg. & $1 \times$ Res. & $2 \times$ Res. & $4 \times$ Res. &  Avg. & $1 \times$ Res. & $2 \times$ Res. & $4 \times$ Res. &  Avg.  \\ \hline
    SARO-GS~\cite{sarogs}          &                        
                              29.81 &   
                              26.68 &
                              24.66 &
                              27.05 &
                              0.936 &   
                              0.879 &
                              0.839 &
                              0.884 &
                              0.088 &   
                              0.179 &
                              0.266 &
                              0.178 \\
Spacetime-GS~\cite{spacetimegaussians}                            & 
                              28.53 &   
                              26.22 &
                              24.43 &
                              26.39 &
                              0.933 &   
                              0.881 &
                              0.846 &
                              0.886 &
                              0.091 &   
                              0.186 &
                              0.267 &
                              0.181 \\
4DGS~\cite{wu20244d}                                &                       
                              28.94 &   
                              25.88 &
                              24.0 &
                              26.28 &
                              0.928 &   
                              0.869 &
                              0.844 &
                              0.880 &
                              0.097 &   
                              0.194 &
                              0.261 &
                              0.184 \\
4DRotorGS~\cite{duan20244d}                          &                       
                              29.64 &   
                              27.26 &
                              27.03 &
                              27.97 &
                              0.939 &   
                              0.885 &
                              0.899 &
                              0.907 &
                              0.082 &   
                              0.189 &
                              0.207 &
                              0.159 \\
Grid4D~\cite{grid4d}   &                       
                              28.88 &   
                              26.03 &
                              24.37 &
                              26.42 &
                              0.915 &   
                              0.856 &
                              0.832 &
                              0.867 &
                              0.108 &   
                              0.207 &
                              0.276 &
                              0.196 \\
Ours   &                       
                              29.56 &   
                              28.17 &
                              27.34 &
                              28.35 &
                              0.934 &   
                              0.898 &
                              0.869 &
                              0.900 &
                              0.092 &   
                              0.175 &
                              0.237 &
                              0.168 
    \end{tabular}
    }
      \label{tab:avg_plenoptic_single_train_multi_test_coffeemartini}
\end{table}

\begin{table}[!t]
    \renewcommand{\tabcolsep}{1pt}
    \centering
    \caption{
    \textbf{Single-Scale Training and Multi-Scale Evaluation on cut roasted beef scene of Plenoptic Video Dataset~\cite{li2022neural}.} All methods are trained at the smallest scale ($1 \times$) that is $676$x$507$ and evaluated across multiple scales ($1 \times$, $2 \times$, and $4 \times$), where increasing resolution simulates zoom-in effects. While our approach performs comparably to existing methods at the training resolution($1 \times$), it demonstrates superior fidelity across all other scales, significantly outperforming prior work.}
    \resizebox{0.95\linewidth}{!}{
    \begin{tabular}{@{}l@{\,\,}|cccc|cccc|cccc}
    \toprule
    & \multicolumn{4}{c|}{PSNR $\uparrow$} & \multicolumn{4}{c|}{SSIM $\uparrow$} & \multicolumn{4}{c}{LPIPS $\downarrow$}  \\
    & $1 \times$ Res. & $2 \times$ Res. & $4 \times$ Res.  & Avg. & $1 \times$ Res. & $2 \times$ Res. & $4 \times$ Res. &  Avg. & $1 \times$ Res. & $2 \times$ Res. & $4 \times$ Res. &  Avg.  \\ \hline
    SARO-GS~\cite{sarogs}          &                        
                              34.19 &   
                              30.34 &
                              28.20 &
                              30.91 &
                              0.964 &   
                              0.928 &
                              0.903 &
                              0.931 &
                              0.084 &   
                              0.157 &
                              0.224 &
                              0.155 \\
Spacetime-GS~\cite{spacetimegaussians}                            & 
                              34.0 &   
                              29.97 &
                              27.75 &
                              30.57 &
                              0.966 &   
                              0.931 &
                              0.909 &
                              0.935 &
                              0.088 &   
                              0.164 &
                              0.224 &
                              0.158 \\
4DGS~\cite{wu20244d}                                &                       
                              31.74 &   
                              29.03 &
                              27.09 &
                              29.28 &
                              0.944 &   
                              0.908 &
                              0.890 &
                              0.914 &
                              0.103 &   
                              0.178 &
                              0.223 &
                              0.168 \\
4DRotorGS~\cite{duan20244d}                          &                       
                              33.15 &   
                              30.44 &
                              30.17 &
                              31.25 &
                              0.949 &   
                              0.914 &
                              0.917 &
                              0.926 &
                              0.109 &   
                              0.184 &
                              0.207 &
                              0.167 \\
Grid4D~\cite{grid4d}   &                       
                              33.64 &   
                              29.75 &
                              28.01 &
                              30.46 &
                              0.957 &   
                              0.915 &
                              0.896 &
                              0.922 &
                              0.087 &   
                              0.170 &
                              0.223 &
                              0.160 \\
Ours   &                       
                              33.32 &   
                              31.97 &
                              31.21 &
                              32.17 &
                              0.955 &   
                              0.937 &
                              0.925 &
                              0.939 &
                              0.095 &   
                              0.164 &
                              0.231 &
                              0.163 
    \end{tabular}
    }
    \vspace{-0.1in}
       \label{tab:avg_plenoptic_single_train_multi_test_cutroastedbeef}
\end{table}

\begin{table}[!t]
    \renewcommand{\tabcolsep}{1pt}
    \caption{
    \textbf{Single-Scale Training and Multi-Scale Evaluation on flame salmon scene of Plenoptic Video Dataset~\cite{li2022neural}.} All methods are trained at the smallest scale ($1 \times$) that is $676$x$507$ and evaluated across multiple scales ($1 \times$, $2 \times$, and $4 \times$), where increasing resolution simulates zoom-in effects. While our approach performs comparably to existing methods at the training resolution($1 \times$), it demonstrates superior fidelity across all other scales, significantly outperforming prior work.}
    \centering
    \resizebox{0.95\linewidth}{!}{
    \begin{tabular}{@{}l@{\,\,}|cccc|cccc|cccc}
    \toprule
    & \multicolumn{4}{c|}{PSNR $\uparrow$} & \multicolumn{4}{c|}{SSIM $\uparrow$} & \multicolumn{4}{c}{LPIPS $\downarrow$}  \\
    & $1 \times$ Res. & $2 \times$ Res. & $4 \times$ Res.  & Avg. & $1 \times$ Res. & $2 \times$ Res. & $4 \times$ Res. &  Avg. & $1 \times$ Res. & $2 \times$ Res. & $4 \times$ Res. &  Avg.  \\ \hline
    SARO-GS~\cite{sarogs}          &                        
                              29.78 &   
                              26.64 &
                              24.68 &
                              27.03 &
                              0.936 &   
                              0.877 &
                              0.836 &
                              0.883 &
                              0.082 &   
                              0.175 &
                              0.264 &
                              0.173 \\
Spacetime-GS~\cite{spacetimegaussians}                            & 
                              29.84 &   
                              26.91 &
                              25.01 &
                              27.25 &
                              0.937 &   
                              0.885 &
                              0.851 &
                              0.891 &
                              0.083 &   
                              0.177 &
                              0.259 &
                              0.173 \\
4DGS~\cite{wu20244d}                                &                       
                              30.38 &   
                              26.35 &
                              24.26 &
                              26.99 &
                              0.932 &   
                              0.870 &
                              0.844 &
                              0.882 &
                              0.089 &   
                              0.188 &
                              0.256 &
                              0.177 \\
4DRotorGS~\cite{duan20244d}                          &                       
                              29.90 &   
                              26.54 &
                              26.60 &
                              27.68 &
                              0.931 &   
                              0.867 &
                              0.805 &
                              0.867 &
                              0.091 &   
                              0.193 &
                              0.263 &
                              0.182 \\
Grid4D~\cite{grid4d}   &                       
                              29.91 &   
                              26.43 &
                              24.65 &
                              26.99 &
                              0.928 &   
                              0.864 &
                              0.837 &
                              0.876 &
                              0.093 &   
                              0.193 &
                              0.264 &
                              0.183 \\
Ours   &                       
                              29.04 &   
                              27.68 &
                              26.81 &
                              27.84 &
                              0.932 &   
                              0.896 &
                              0.864 &
                              0.897 &
                              0.084 &   
                              0.168 &
                              0.244 &
                              0.165 
    \end{tabular}
    }
       \label{tab:avg_plenoptic_single_train_multi_test_flamesalmon}
\end{table}

\begin{table}[!t]
    \renewcommand{\tabcolsep}{1pt}
    \centering
    \caption{
    \textbf{Single-Scale Training and Multi-Scale Evaluation on flame steak scene of Plenoptic Video Dataset~\cite{li2022neural}.} All methods are trained at the smallest scale ($1 \times$) that is $676$x$507$ and evaluated across multiple scales ($1 \times$, $2 \times$, and $4 \times$), where increasing resolution simulates zoom-in effects. While our approach performs comparably to existing methods at the training resolution($1 \times$), it demonstrates superior fidelity across all other scales, significantly outperforming prior work.}
    \resizebox{0.95\linewidth}{!}{
    \begin{tabular}{@{}l@{\,\,}|cccc|cccc|cccc}
    \toprule
    & \multicolumn{4}{c|}{PSNR $\uparrow$} & \multicolumn{4}{c|}{SSIM $\uparrow$} & \multicolumn{4}{c}{LPIPS $\downarrow$}  \\
    & $1 \times$ Res. & $2 \times$ Res. & $4 \times$ Res.  & Avg. & $1 \times$ Res. & $2 \times$ Res. & $4 \times$ Res. &  Avg. & $1 \times$ Res. & $2 \times$ Res. & $4 \times$ Res. &  Avg.  \\ \hline
    SARO-GS~\cite{sarogs}          &                        
                              34.28 &   
                              31.17 &
                              29.26 &
                              31.57 &
                              0.968 &   
                              0.933 &
                              0.911 &
                              0.937 &
                              0.086 &   
                              0.148 &
                              0.210 &
                              0.148 \\
Spacetime-GS~\cite{spacetimegaussians}                            & 
                              33.79 &   
                              30.78 &
                              28.69 &
                              31.08 &
                              0.971 &   
                              0.940 &
                              0.919 &
                              0.943 &
                              0.087 &   
                              0.155 &
                              0.207 &
                              0.149 \\
4DGS~\cite{wu20244d}                                &                       
                              30.73 &   
                              28.66 &
                              27.41 &
                              28.93 &
                              0.954 &   
                              0.919 &
                              0.902 &
                              0.925 &
                              0.096 &   
                              0.164 &
                              0.209 &
                              0.156 \\
4DRotorGS~\cite{duan20244d}                          &                       
                              33.51 &   
                              31.12 &
                              29.95 &
                              31.52 &
                              0.969 &   
                              0.930 &
                              0.907 &
                              0.935 &
                              0.071 &   
                              0.171 &
                              0.202 &
                              0.148 \\
Grid4D~\cite{grid4d}   &                       
                              29.89 &   
                              28.21 &
                              27.21 &
                              28.43 &
                              0.950 &   
                              0.914 &
                              0.899 &
                              0.921 &
                              0.082 &   
                              0.158 &
                              0.205 &
                              0.151 \\
Ours   &                       
                              34.08 &   
                              32.20 &
                              30.97 &
                              32.41 &
                              0.966 &   
                              0.938 &
                              0.911 &
                              0.938 &
                              0.084 &   
                              0.159 &
                              0.234 &
                              0.159 
    \end{tabular}
    }
    \vspace{-0.1in}
       \label{tab:avg_plenoptic_single_train_multi_test_flamesteak}
\end{table}

\begin{table}[!t]
    \renewcommand{\tabcolsep}{1pt}
    \centering
      \caption{
    \textbf{Single-Scale Training and Multi-Scale Evaluation on sear steak scene of Plenoptic Video Dataset~\cite{li2022neural}.} All methods are trained at the smallest scale ($1 \times$) that is $676$x$507$ and evaluated across multiple scales ($1 \times$, $2 \times$, and $4 \times$), where increasing resolution simulates zoom-in effects. While our approach performs comparably to existing methods at the training resolution($1 \times$), it demonstrates superior fidelity across all other scales, significantly outperforming prior work.}  
    \resizebox{0.95\linewidth}{!}{
    \begin{tabular}{@{}l@{\,\,}|cccc|cccc|cccc}
    \toprule
    & \multicolumn{4}{c|}{PSNR $\uparrow$} & \multicolumn{4}{c|}{SSIM $\uparrow$} & \multicolumn{4}{c}{LPIPS $\downarrow$}  \\
    & $1 \times$ Res. & $2 \times$ Res. & $4 \times$ Res.  & Avg. & $1 \times$ Res. & $2 \times$ Res. & $4 \times$ Res. &  Avg. & $1 \times$ Res. & $2 \times$ Res. & $4 \times$ Res. &  Avg.  \\ \hline
    SARO-GS~\cite{sarogs}          &                        
                              34.45 &   
                              31.12 &
                              29.25 &
                              31.60 &
                              0.967 &   
                              0.935 &
                              0.914 &
                              0.938 &
                              0.091 &   
                              0.154 &
                              0.217 &
                              0.154 \\
Spacetime-GS~\cite{spacetimegaussians}                            & 
                              34.59 &   
                              31.31 &
                              29.10 &
                              31.66 &
                              0.971 &   
                              0.941 &
                              0.921 &
                              0.944 &
                              0.088 &   
                              0.157 &
                              0.212 &
                              0.152 \\
4DGS~\cite{wu20244d}                                &                       
                              33.33 &   
                              30.14 &
                              27.95 &
                              30.47 &
                              0.962 &   
                              0.926 &
                              0.908 &
                              0.932 &
                              0.092 &   
                              0.162 &
                              0.207 &
                              0.153 \\
4DRotorGS~\cite{duan20244d}                          &                       
                              33.28 &   
                              31.09 &
                              30.06 &
                              31.47 &
                              0.959 &   
                              0.935 &
                              0.918 &
                              0.937 &
                              0.094 &   
                              0.164 &
                              0.219 &
                              0.159 \\
Grid4D~\cite{grid4d}   &                       
                              30.81 &   
                              28.99 &
                              27.91 &
                              29.23 &
                              0.956 &   
                              0.922 &
                              0.909 &
                              0.929 &
                              0.099 &   
                              0.160 &
                              0.207 &
                              0.155 \\
Ours   &                       
                              33.88 &   
                              32.36 &
                              31.42 &
                              32.55 &
                              0.965 &   
                              0.944 &
                              0.926 &
                              0.945 &
                              0.088 &   
                              0.151 &
                              0.218 &
                              0.152 
    \end{tabular}
    }
   \label{tab:avg_plenoptic_single_train_multi_test_searsteak}
\end{table}


\begin{table}[!t]
    \renewcommand{\tabcolsep}{1pt}
    \centering
    \caption{
    \textbf{Temporal Stability Evaluation on sear steak scene of Plenoptic Video Dataset~\cite{li2022neural}.} All methods are trained at the smallest scale ($1 \times$) that is $676$x$507$ and evaluated across multiple scales ($1 \times$, $2 \times$, and $4 \times$), where increasing resolution simulates zoom-in effects. } 
    \resizebox{\linewidth}{!}{
    \begin{tabular}{@{}l@{\,\,}|cccc|cccc|cccc}
    \toprule
    & \multicolumn{4}{c|}{\textbf{W-Temporal Error} $\downarrow$} & \multicolumn{4}{c|}{\textbf{W-LPIPS} $\downarrow$} & \multicolumn{4}{c}{\textbf{W-SSIM} $\uparrow$}  \\
    & $1 \times$ Res. & $2 \times$ Res. & $4 \times$ Res.  & Avg. & $1 \times$ Res. & $2 \times$ Res. & $4 \times$ Res. &  Avg. & $1 \times$ Res. & $2 \times$ Res. & $4 \times$ Res. &  Avg.  \\ \hline
    Spacetime-GS~\cite{spacetimegaussians}                            & 
                                 0.902 & 0.903 & 0.848 & 0.885 & 0.002952 & 0.002609 & 0.002557 & 0.002706 & 0.992751 & 0.992130 & 0.991626 & 0.992169 \\
4DGS~\cite{wu20244d}                                &                       
                              0.622 & 0.573 & 0.518 & 0.571 & 0.002215 & 0.001695 & 0.001372 & 0.001761 & 0.994050 & 0.994380 & 0.994196 & 0.994209 \\
4DRotorGS~\cite{duan20244d}                          &                       
                              0.687 & 0.632 & 0.569 & 0.629 & 0.002014 & 0.001372 & 0.002215 & 0.001695 & 0.994316 & 0.994404 & 0.994534 & 0.994418 \\
Grid4D~\cite{grid4d}   &                       
                              0.700 & 0.696 & 0.646 & 0.681 & 0.002480 & 0.001954 & 0.001673 & 0.002036 & 0.993439 & 0.993699 & 0.993482 & 0.993540 \\
Ours   &                       
                              0.529 & 0.497 & 0.448 & 0.491 & 0.001685 & 0.001213 & 0.000978 & 0.001292 & 0.995878 & 0.995528 & 0.995704 & 0.995703  
    \end{tabular}
    }
    
      \label{tab:temporal_plenoptic_searsteak}
\end{table}

\begin{table}[!t]
    \renewcommand{\tabcolsep}{1pt}
    \centering
    \caption{
    \textbf{Temporal Stability Evaluation on flame steak scene of Plenoptic Video Dataset~\cite{li2022neural}.} All methods are trained at the smallest scale ($1 \times$) that is $676$x$507$ and evaluated across multiple scales ($1 \times$, $2 \times$, and $4 \times$), where increasing resolution simulates zoom-in effects. }
    \resizebox{\linewidth}{!}{
    \begin{tabular}{@{}l@{\,\,}|cccc|cccc|cccc}
    \toprule
    & \multicolumn{4}{c|}{\textbf{W-Temporal Error} $\downarrow$} & \multicolumn{4}{c|}{\textbf{W-LPIPS} $\downarrow$} & \multicolumn{4}{c}{\textbf{W-SSIM} $\uparrow$}  \\
    & $1 \times$ Res. & $2 \times$ Res. & $4 \times$ Res.  & Avg. & $1 \times$ Res. & $2 \times$ Res. & $4 \times$ Res. &  Avg. & $1 \times$ Res. & $2 \times$ Res. & $4 \times$ Res. &  Avg.  \\ \hline
    
Spacetime-GS~\cite{spacetimegaussians}                            & 
                                 1.198 & 1.184 & 1.120 & 1.167 & 0.004824 & 0.004476 & 0.004407 & 0.004569 & 0.990170 & 0.989600 & 0.989086 & 0.989619 \\
4DGS~\cite{wu20244d}                                &                       
                              1.083 & 1.033 & 1.005 & 1.041 & 0.003779 & 0.003297 & 0.003217 & 0.003431 & 0.991249 & 0.990875 & 0.990234 & 0.990786 \\ 
4DRotorGS~\cite{duan20244d}                          &                       
                             0.966 & 0.877 & 0.800 & 0.881 & 0.003855 & 0.003247 & 0.002917 & 0.003340 & 0.991956 & 0.992280 & 0.992499 & 0.992245 \\
Grid4D~\cite{grid4d}   &                       
                              0.862 & 0.823 & 0.778 & 0.821 & 0.003471 & 0.002891 & 0.002631 & 0.002998 & 0.992322 & 0.992116 & 0.992019 & 0.992152 \\
Ours   &                       
                              0.876 & 0.832 & 0.776 & 0.828 & 0.003680 & 0.003232 & 0.003001 & 0.003304 & 0.992189 & 0.992245 & 0.992340 & 0.992258 
    \end{tabular}
    }
       \label{tab:temporal_plenoptic_flameteak}
\end{table}

\begin{table}[!t]
    \renewcommand{\tabcolsep}{1pt}
    \centering
    \caption{
    \textbf{Temporal Stability Evaluation on flame salmon scene of Plenoptic Video Dataset~\cite{li2022neural}.} All methods are trained at the smallest scale ($1 \times$) that is $676$x$507$ and evaluated across multiple scales ($1 \times$, $2 \times$, and $4 \times$), where increasing resolution simulates zoom-in effects. }
    \resizebox{\linewidth}{!}{
    \begin{tabular}{@{}l@{\,\,}|cccc|cccc|cccc}
    \toprule
    & \multicolumn{4}{c|}{\textbf{W-Temporal Error} $\downarrow$} & \multicolumn{4}{c|}{\textbf{W-LPIPS} $\downarrow$} & \multicolumn{4}{c}{\textbf{W-SSIM} $\uparrow$}  \\
    & $1 \times$ Res. & $2 \times$ Res. & $4 \times$ Res.  & Avg. & $1 \times$ Res. & $2 \times$ Res. & $4 \times$ Res. &  Avg. & $1 \times$ Res. & $2 \times$ Res. & $4 \times$ Res. &  Avg.  \\ \hline
    
Spacetime-GS~\cite{spacetimegaussians}                            & 
                                 1.265 & 1.281 & 1.220 & 1.255  & 0.004824 & 0.004476 & 0.004407 & 0.004569 &  0.991684 & 0.991092 & 0.990413 & 0.991063 \\
4DGS~\cite{wu20244d}                                &                       
                             0.853 & 0.791 & 0.707 & 0.784  & 0.003779 & 0.003297 & 0.003217 & 0.003431 &  0.993425 & 0.993351 & 0.993226 & 0.993334 \\  
4DRotorGS~\cite{duan20244d}                          &                       
                             0.929 & 0.871 & 0.803 & 0.868  & 0.003855 & 0.003247 & 0.002917 & 0.003340 &  0.993411 & 0.993640 & 0.993664 & 0.993572 \\
Grid4D~\cite{grid4d}   &                       
                              0.941 & 0.954 & 0.913 & 0.936  & 0.003471 & 0.002891 & 0.002631 & 0.002998 &  0.993060 & 0.992848 & 0.992499 & 0.992802 \\ 
Ours   &                       
                              0.606 & 0.607 & 0.574 & 0.596  & 0.003680 & 0.003232 & 0.003001 & 0.003305 &  0.995794 & 0.995932 & 0.995966 & 0.995897  
    \end{tabular}
    }

       \label{tab:temporal_plenoptic_flamesalmon}
\end{table}

\begin{figure}[!t]
    \centering
    \includegraphics[width=0.9\linewidth]{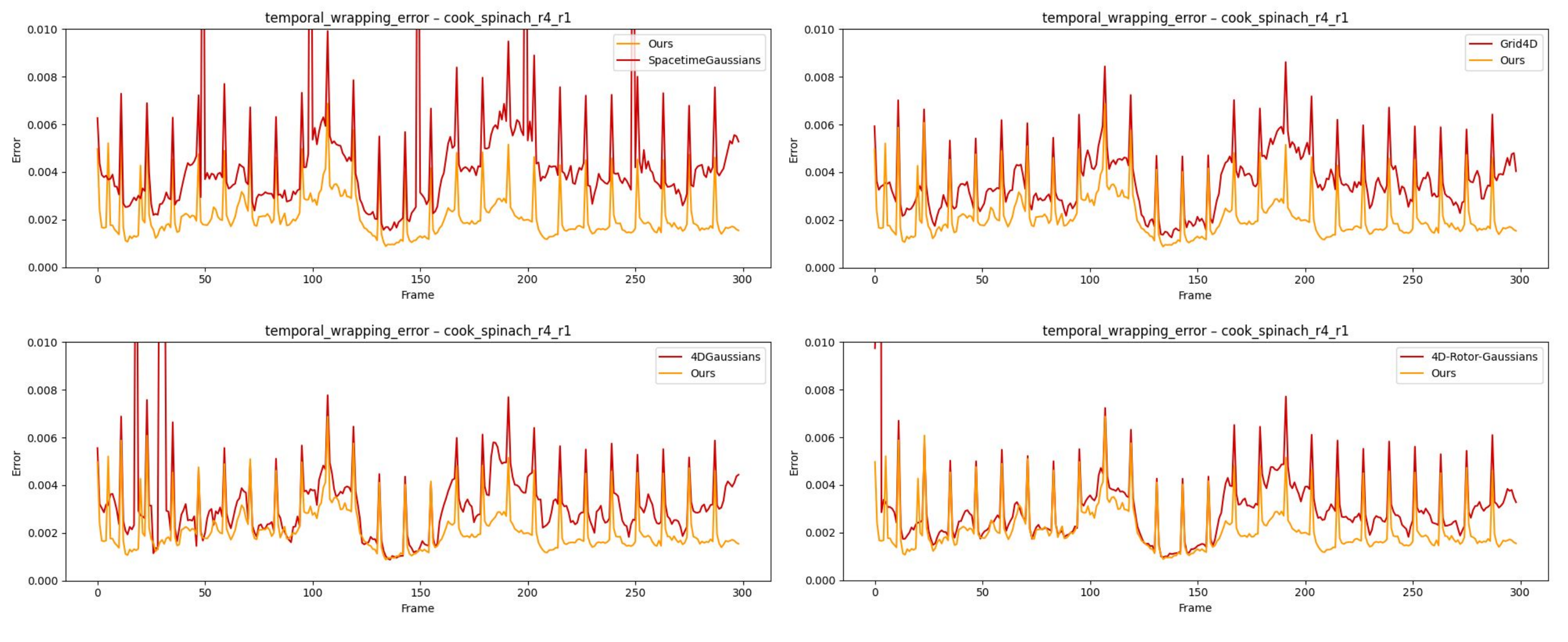}
    \caption{Here we show temporal plot of warping error for cook spinach scene which is trained at the smallest scale ($1 \times$) that is $676$x$507$ and evaluated across ($4 \times$) rendered novel-views. The proposed method consistently exhibits a lower error than the baseline methods, indicating superior temporal stability.}
    \label{fig:supp_temporal_cook_spinach}
\end{figure}

\end{document}